\documentclass{article} 
\usepackage{iclr2023_conference,times}

\usepackage{amsmath,amsfonts,bm}

\def\eqref#1{equation~\ref{#1}}

\def\1{\bm{1}}

\DeclareMathAlphabet{\mathsfit}{\encodingdefault}{\sfdefault}{m}{sl}
\SetMathAlphabet{\mathsfit}{bold}{\encodingdefault}{\sfdefault}{bx}{n}

\usepackage{hyperref}
\usepackage{url}

\usepackage{booktabs}       
\usepackage{amsfonts}       
\usepackage{nicefrac}       
\usepackage{microtype}      
\usepackage{xcolor}         

\usepackage{microtype}
\usepackage{graphicx}

\usepackage{subcaption}

\usepackage{amsmath}
\usepackage[capitalise]{cleveref} 
\usepackage{amssymb}
\usepackage{units}
\usepackage{multirow}
\usepackage{xr}
\usepackage{xfrac}
\usepackage{mathtools}
\usepackage{algpseudocode}

\title{Accurate Neural Training with 4-bit Matrix Multiplications at Standard Formats}

\author{
\textbf{Brian Chmiel\,${^\dagger}{^\circ}$}\quad
\textbf{Ron Banner\,$^\dagger$}\quad
\textbf{ Elad Hoffer\,${^\dagger}$}\quad
\textbf{ Hilla Ben Yaacov\,${^\dagger}$}\quad
\textbf{ Daniel Soudry\,$^\circ$}
\\[0.2cm]
$^\dagger$Habana Labs  --  An Intel company, Caesarea, Israel,\\
$^\circ$Department of Electrical Engineering - Technion, Haifa, Israel
\\[0.2cm]
\small{\texttt{\{\href{mailto:bchmiel@habana.ai}{bchmiel}, \href{mailto:rbanner@habana.ai}{rbanner}, \href{mailto:ehoffer@habana.ai}{ehoffer}, \href{mailto:hbyaacov@habana.ai}{hbyaacov}\}@habana.ai}}\\
\small{\texttt{\{\href{mailto:daniel.soudry@gmail.com}{daniel.soudry}\}@gmail.com}}\\
}

\iclrfinalcopy 
\begin{document}

\newcommand\RB[1]{\textcolor{blue}{(\textbf{Ron}: #1)}} 
\newcommand\BCH[1]{\textcolor{green}{(\textbf{Brian}: #1)}} 
\newcommand\DS[1]{\textcolor{cyan}{(\textbf{Daniel}: #1)}} 
\newcommand\EH[1]{\textcolor{red}{(\textbf{Elad}: #1)}} 

 \newcommand\rebuttal[1]{\textcolor{black}{#1}}

\maketitle

\begin{abstract}
Quantization of the weights and activations is one of the main methods to reduce the computational footprint of Deep Neural Networks (DNNs) training. Current methods enable 4-bit quantization of the forward phase. However, this constitutes only a third of the training process. Reducing the computational footprint of the entire training process requires the quantization of the neural gradients, i.e., the loss gradients with respect to the outputs of intermediate neural layers. 

Previous works separately showed that accurate 4-bit quantization of the neural gradients needs to (1) be unbiased and (2) have a log scale. However, no previous work aimed to combine both ideas, as we do in this work. Specifically, we examine the importance of having unbiased quantization in quantized neural network training, where to maintain it, and how to combine it with logarithmic quantization. Based on this, we suggest a \textit{logarithmic unbiased quantization} (LUQ) method to quantize both the forward and backward phases to 4-bit, achieving state-of-the-art results in 4-bit training without the overhead. For example, in ResNet50 on ImageNet, we achieved a degradation of 1.1\%. We further improve this to a degradation of only 0.32\% after three epochs of high precision fine-tuning, combined with a variance reduction method---where both these methods add overhead comparable to previously suggested methods. 
\end{abstract}

\section{Introduction}

 Deep neural networks (DNNs) training consists of three main general-matrix-multiply (GEMM) phases: the forward phase, backward phase, and update phase. Quantization has become one of the main methods to compress DNNs and reduce the GEMM computational resources. Previous works showed the weights and activations in the forward pass to 4 bits while preserving model accuracy \citep{banner2018post,Nahshan2019LossAP,Bhalgat2020LSQIL,PACT}. Despite these advances, they only apply to a third of the training process, while the backward phase and update phase are still computed with higher precision. 

Recently, \cite{UltraLowP4} was able, for the first time, to train a DNN while reducing the numerical precision of most of its parts to 4 bits with some degradation (e.g., $2.49\%$ error in ResNet50). To do so,  \citet{UltraLowP4} suggested a non-standard radix-4 floating-point format, combined with double quantization of the neural gradients (called two-phase rounding). This was an impressive step forward in the ability to quantize all GEMMs in training. However, since a radix-4 format is not aligned with conventional radix-2, any numerical conversion between the two requires an explicit multiplication to modify both the exponent and mantissa. Thus, their non-standard quantization requires specific hardware support \citep{Kupriianova_2013} that can significantly reduce the benefit of quantization to low bits (\cref{sec:UltraLowCompare}), and make it less practical.

The main challenge in reducing the numerical precision of the entire training process is quantizing the neural gradients, i.e. the backpropagated error. Previous works showed separately that, to achieve accurate low precision representation of the neural gradients, it is important to use: (1) Logarithmic quantization and (2) Unbiased quantization. Specifically, \citet{Chmiel2020NeuralGA} showed the neural gradients have a heavy tailed near-lognormal distribution and found an analytical expression for the optimal floating point format. At low precision levels, the optimal format is logarithmically quantized. For example, for FP4 the optimal format is [sign,exponent,mantissa] = [1,3,0],  i.e. without mantissa bits. In contrast, weights and activations are well approximated with  Normal or Laplacian distributions \citep{banner2018post,SAWB}, and therefore are better approximated using uniform quantization (e.g., INT4). However, \citet{Chmiel2020NeuralGA} did not use unbiased quantization (nor did any of the previous works that use logarithmic quantization of the neural gradients \citep{Li2020AdditivePQ,Miyashita2016ConvolutionalNN,Ortiz2018LowPrecisionFS}). Therefore, they were unable to successfully quantize in this FP4 format (their narrowest format was FP5). 

\citet{chen2020statistical} showed that unbiased quantization of the neural gradients is essential to get unbiasedness in the weight gradients, which is required in SGD analysis convergence \citep{bottou2018optimization}. However, they focused on quantization using integer formats, as did other works that pointed out on the importance of being unbiased \citep{Banner2018ScalableMF,Zhong2022ExploringTP}.
Naive quantization of the neural gradients using the optimal FP4 format (logarithmic) results in biased estimates of the FP32 weight gradients---and this leads to severe degradation in the test accuracy. For example, a major issue is that under aggressive (naive) quantization many neural gradients with magnitudes below the representable range are zeroed, resulting in biased estimates of the FP32 gradients and reduced model accuracy. 

Using either a logarithmic scale or unbiased rounding alone catastrophically fails at 4bit quantization of the neural gradients (e.g., see \cref{fig:repetitionsAndAblations} below). Therefore, it is critical to combine them, as we do in this paper in \cref{sec:luq}. To do this, we stochastically quantize gradients below the representable range to either zero or the smallest representable magnitude $\alpha$ to provide unbiased estimates within that "underflow" range. Additionally, in order to represent the maximum magnitude without bias, we dynamically adjust $\alpha$ so that the maximum can always be represented with an exponentiated scaling starting at $\alpha$. Finally, to completely eliminate bias, we devise an efficient way to use stochastic rounding on a logarithmic scale, on the values between $\alpha$ and the maximum. Together, this gradient quantization method is called \textit{Logarithmic Unbiased Quantization (LUQ)}, and for 4-bit quantization it uses a numerical format with one sign bit, three exponent bits, and zero mantissa bits, along with stochastic mapping (to zero or $\alpha$) of gradients whose values are below $\alpha$ and stochastic rounding within the representable range.

\textbf{Main contribution} LUQ, for the first time, combines logarithmic quantization with unbiased quantization for the neural gradients and does this efficiently using a standard format. By additionally quantizing the forward phase to INT4, we enable, for the first time, an efficient scheme for ``full 4-bit training", i.e. the weights, activations and neural gradients are quantized to 4-bit in standard formats (see \cref{sec:AppQFormats}) so all GEMMs can be done in 4-bit, and also bandwidth can be reduced. As we show, this method requires little to no overhead while achieving state-of-the-art accuracy results: for example, in ResNet50 we get $1.1\%$ error degradation with standard formats; in comparison, the previous method \citep{UltraLowP4} had $2.49\%$ error degradation but required non-standard formats, as well as additional modifications which have additional overhead. 

Moreover, in \cref{sec:optional methods} we suggest two optional simple methods to further reduce the degradation, with some overhead: the first method reduces the quantization variance of the neural gradients using re-sampling, while the second is fine-tuning in high precision. Combining LUQ with these two proposed methods we achieve, for the first time, only $0.32\%$ error in the 4-bit training of ResNet50. The overhead of our additional methods is no more than similar modifications previously suggested in \cite{UltraLowP4}.  Lastly, in \cref{sec:discussion} we discuss how to reduce remaining overheads such as data movement, scaling operations, and GEMM-related operations. 


    

\section{Related works}
Neural networks Quantization has been extensively investigated in the last few years. Most of the quantization research has focused on reducing the numerical precision of the weights and activations for inference (e.g., \cite{BNN,XnotNet,banner2018post,Nahshan2019LossAP,PACT,Bhalgat2020LSQIL,SAWB,Liang2021PruningAQ}). In this case, for standard ImageNet models, the best performing methods can achieve quantization to 4 bits with small or no degradation \cite{SAWB,Sakr2022OptimalCA}. These methods can be used to reduce the computational resources in approximately a third of the training (\cref{eq:forward}). However, without quantizing the neural gradients, we cannot reduce the computational resources in the remaining two thirds of the training process (\cref{eq:backward} and \cref{eq:update}). An orthogonal approach is low precision for the gradients of the weights in distributed training \citep{Alistarh2016QSGDCS,Bernstein2018signSGDCO} in order to reduce the bandwidth and not the training computational resources. 

\citet{Sakr2019PerTensorFQ} suggest a systematic approach to design a full training using fixed point quantization which includes mixed-precision quantization. \citet{Banner2018ScalableMF} first showed that it is possible to use INT8 quantization for the weights, activations, and neural gradients, thus reducing the computational footprint of most parts of the training process. Concurrently, \cite{Wang2018TrainingDN} was the first work to achieve full training in FP8 format. Additionally, they suggested a method to reduce the accumulator precision from 32bit to 16 bits, by using chunk-based accumulation and floating point stochastic rounding. Later, \citet{Wiedemann2020DitheredBA} showed full training in INT8 with improved convergence, by applying a stochastic quantization scheme to the neural gradients called non-subtractive-dithering (NSD).  Also, \cite{Sun2019Hybrid8F} presented a novel hybrid format for full training in FP8, while the weights and activations are quantized to  [1,4,3] format, the neural gradients are quantized to [1,5,2] format to catch a wider dynamic range. \citet{Fournarakis2021InHindsightQR} suggested a method to reduce the data traffic during the calculation of the quantization range, using the moving average of the tensor's statistics.

While it appears that it is possible to quantize to 8-bits all computational elements in the training process,  4-bits quantization of the neural gradients is still challenging. \citet{Chmiel2020NeuralGA} suggested that this difficulty stems from the heavy-tailed distribution of the neural gradients, which can be approximated with a lognormal distribution. This distribution is more challenging to quantize in comparison to the normal distribution which is usually used to approximate the weights or activations \citep{banner2018post}. Different works \citep{Li2020AdditivePQ,Miyashita2016ConvolutionalNN} tried to use logarithmic quantization for the neural gradients, however, they failed to quantize them unbiasedly. 

\citet{UltraLowP4} was the first work that presented a method to reduce the numerical precision to 4-bits for the vast majority of the computations needed during DNNs training. They use known methods to quantize the forward phase to INT4 and suggested quantizing the neural gradients twice with a non-standard radix-4 FP4 format. The use of the radix-4, instead of the commonly used radix-2 format, allows for covering a wider dynamic range. The main problem with their method is the specific hardware support for their suggested radix-4 datatype, which may limit the practicality of implementing their suggested data type. 

\citet{Chen2020ASF} suggested reducing the variance in neural gradients quantization by dividing them into several blocks and quantizing each to INT4 separately. They require expensive sorting. Additionally, their per-sample quantization do not allow the use of an efficient GEMM operation. 

\section{Background: Quantization formats and rounding schemes}

Which quantization schemes should we use in 4bit training? Previous works \citep{SAWB,banner2018post} showed that weights and activations can be quantized to INT4 with little to no accuracy degradation. In contrast, for the neural gradients, a recent work \citep{Chmiel2020NeuralGA} showed analytically that the optimal format is logarithmic ([1,3,0]). Combining all these schemes, we focus on full 4-bit training using standard formats, with the following three 4-bit quantized GEMMs:
\begin{equation}
    \textbf{[Forward]} \quad z_{l} =  Q_{\mathrm{INT}}(W_l) Q_{\mathrm{INT}}(a_{l-1}); \quad
    \quad a_l = f_l(z_{l})
    \label{eq:forward}
\end{equation}
\vspace{-2mm}
\begin{equation}
    \textbf{[Backward]} \quad g_{l-1}  =  Q_{\mathrm{INT}}(W_l^T)  Q_{\mathrm{FP}}(\delta_{l}); \quad  \delta_{l} = f_l^{\prime}(z_{l}) \odot g_l 
    \label{eq:backward}
\end{equation}
\vspace{-2mm}
\begin{equation}
\label{eq:update}
    \textbf{[Update]} \quad \frac{\partial C}{\partial W_{l}} = Q_{\mathrm{FP}}(\delta_{l}) Q_{\mathrm{INT}}(a_{l-1}^{T}) \,,
\end{equation}
where  $C$ is the loss function,  $\odot$ is a component-wise product and, in each layer $l$,  $f_l$ is the activation function, the weights ($W_l$) and activations ($a_l$) are quantized with INT4 ($Q_{\mathrm{INT}}$) while the neural gradients $\delta_l\triangleq\frac{\partial C}{\partial z_l}$ are quantized with logarithmic FP4 ($Q_{\mathrm{FP}}$), $z_l$ are the pre-activations, and $g_l\triangleq\frac{\partial C}{\partial a_l}$.

Next, we aim to find which rounding scheme should we use in each quantizer. Thus, we study the effects of unbiased rounding during the three phases of training (Eqs. \ref{eq:forward}, \ref{eq:backward}, and \ref{eq:update}). We show that rounding-to-nearest (RDN) should be applied for the weights and activations ($Q_{\mathrm{INT}}$),  while the unbiased method of stochastic rounding (SR) is more suitable for the neural gradients ($Q_{\mathrm{FP}}$).

\subsection{Mean square error comparison} \label{sec: MSE comparison}
In this section, we show that, although stochastic rounding (SR) is unbiased, it generically has a worse mean-square error (MSE) compared to round-to-nearest (RDN). Given that we want to quantize $x$ in a bin with a lower limit $l(x)$ and an upper limit $u(x)$, stochastic rounding can be stated as follows:
\begin{equation}
\label{SR_eq}
\mathrm{SR}(x)= \begin{cases}l(x) , &  w.p. \quad p(x) = 1-\frac{x-l(x)}{u(x)-l(x)}\\ u(x), & w.p. \quad 1-p(x) = \frac{x-l(x)}{u(x)-l(x)}\end{cases} \, .
\end{equation}

The expected rounding value is given by
\begin{equation}
\label{expecration_SR}
\begin{split}
    E[\mathrm{SR}(x)] &= l(x) \cdot p(x) + u(x)\cdot (1-p(x)) = x\, ,
\end{split}
\end{equation}
where here and below the expectation is over the randomness of SR (i.e., $x$ is a deterministic constant).
In \cref{tab:biasVarMse} we present the bias, variance, and MSE of RDN and SR. Full derivatives appear in \cref{sec:appFullMSE}.  

\begin{table*}[h!]
\centering
\caption{Comparison of the bias, variance, and MSE of two different rounding schemes: round-to-nearest (RDN) and stochastic rounding (SR)}
\label{tab:biasVarMse}
\begin{tabular}{lccc}
\toprule
\multicolumn{1}{l|}{Rounding} & \multicolumn{1}{c|}{Bias} & \multicolumn{1}{c|}{Variance} & \multicolumn{1}{c}{MSE}  \\ \midrule
\multicolumn{1}{l|}{RDN} & \multicolumn{1}{c|}{$\min{\left(x - l(x) , u(x) -  x \right)}$} & \multicolumn{1}{c|}{0} & \multicolumn{1}{c}{$\left[\min{\left(x - l(x) , u(x) -  x \right)}\right]^2$}\\ \hline
\multicolumn{1}{l|}{SR} & \multicolumn{1}{c|}{0} & \multicolumn{1}{c|}{$\left(x-l(x)\right)\cdot \left(u(x)-x\right)$} & \multicolumn{1}{c}{$\left(x-l(x)\right)\cdot \left(u(x)-x\right)$} \\ \bottomrule
\end{tabular}

\end{table*}

From \cref{tab:biasVarMse}, since $\min(a,b)^2\leq a\cdot b$ for every $a,b$, we have that\footnote{Note \cref{inequality} implies that $\int p(x) \mathrm{MSE}[SR(x)] dx \geq \int p(x) \mathrm{MSE}[RDN(x)] dx$ for any distribution $p(x)$. } 
\begin{equation}
\text{MSE}\left[SR\left( x \right)\right] \geq \text{MSE}[RDN(x)],  \quad  \forall x \, .
\label{inequality}
\end{equation}

In \cref{fig:msrSrRnd} we plot the mean-square-error for $x\in [0,1]$, $l(x)=0$, and $u(x)=1$. However, while round-to-nearest has a lower MSE than SR, the former is a biased estimator. 

\subsection{When is it important to use unbiased quantization?}\label{sub: unbiased gradients}
To prove convergence, textbook analyses of SGD typically assume the expectation of the (mini-batch) weight gradients is sufficiently close to the true (full-batch) gradient (e.g., assumption 4.3 in \citep{bottou2018optimization}). This assumption is satisfied when the weight gradients are unbiased. Next, we explain (as pointed out by previous works, such as \cite{chen2020statistical}) that the weight gradients are unbiased when the neural gradients are quantized stochastically without bias. 

Recall $W_l$ and  $f_l$ are, respectively, the weights and activation function at layer $l$ and $C$ is the cost function. Given an input–output pair $(x,y)$, the loss is:
\begin{equation}
\label{fwd}
C\left(y, f_{L}\left(W_{L} f_{L-1}\left(W_{L-1} \cdots f_{2}\left(W_{2} f_{1}\left(W_{1} x\right)\right) \cdots\right)\right)\right) \, .
\end{equation}

\textbf{Backward and Update Phases}  Recall $z_{l}$ and $a_{l}$ are, respectively, the pre- and post- activations of layer $l$, and $\delta_{l}\triangleq\frac{dC}{dz_{l}}$. Defining $\delta_{l}^q\triangleq Q_{\mathrm{FP}}(\delta_{l})$, we show in \cref{sec:appFullUnbias} that
the gradient of the weights in layer $l$ is $\nabla_{W_{l}} C=\delta_{l}a_{l-1}^{\top}$ and its quantized form is $\nabla_{W_{l}} C_q=\delta_{l}^q  a_{l-1}^{\top}$. Therefore, the update $\nabla_{W_{l}} C_q$ is an unbiased estimator of $\nabla_{W_{l}} C$:
\begin{equation}
\begin{split}
\label{update}
    E\left[\nabla_{W_{l}} C_q\right]&=E\left[\delta_{l}^q  a_{l-1}^{\top}\right] = E\left[\delta_{l}^q\right]a_{l-1}^{\top} = \delta_{l} a_{l-1}^{\top} =  \nabla_{W_{l}} C \, .
\end{split}
\end{equation}

\textbf{Forward phase} The forward phase is different from the backward and updates phases in that unbiasedness at the tensor level is not necessarily a guarantee of unbiasedness at the model level since the activation functions and loss functions are not linear. For example, suppose we have two weight layers $W_1, W_2$, activation $f$, input $x$, and an SR quantizer $Q$. Then, despite that $Q$ is unbiased (i.e., $EQ(x)=x$), we get:
\begin{equation}
 \mathbb{E}[ f (W_2 Q (f ( W_1 x))]  \neq  \mathbb{E}[ f (W_2  (f ( W_1 x))] 
 \label{eq:forwardbias}
 \end{equation}
 since $f$ is non-linear.
 This means there is no point to use SR in the forward path since it will increase the MSE (\cref{inequality}), but it will not fix the bias issue. 

\subsection{Conclusions: when to use each rounding scheme?} 
\label{sec:conclusionsFormats}
 Following the above results, the activation and weights quantization in the forward phase ($Q_{INT}$ in eq. \cref{eq:forward}) should use RDN. This is because SR will increase the MSE (as shown in \cref{inequality}), an increase which typically harms the final accuracy\footnote{There are some cases where adding limited noise, such as dropout, locally increases the MSE but improves generalization. However, this is typically not the case, especially if the noise is large.}, but will not help make the loss estimate unbiased, due to the non-linearity of the loss and activation functions (e.g., \cref{eq:forwardbias}). To avoid mismatch (and additional bias), we use RDN in $Q_{\mathrm{INT}}$ also in the other phases of training.

\begin{figure*}[h]
\begin{subfigure}{0.33\textwidth}
  \centering
  \includegraphics[width=\linewidth]{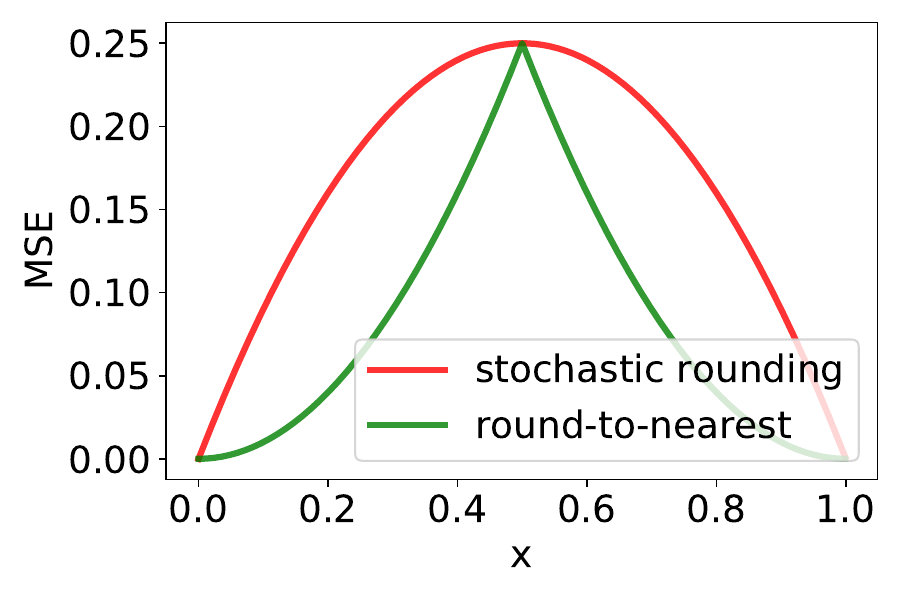}  
  \caption{}
   \label{fig:msrSrRnd}
 \end{subfigure}
\begin{subfigure}{0.33\textwidth}
  \centering
  \includegraphics[width=\linewidth]{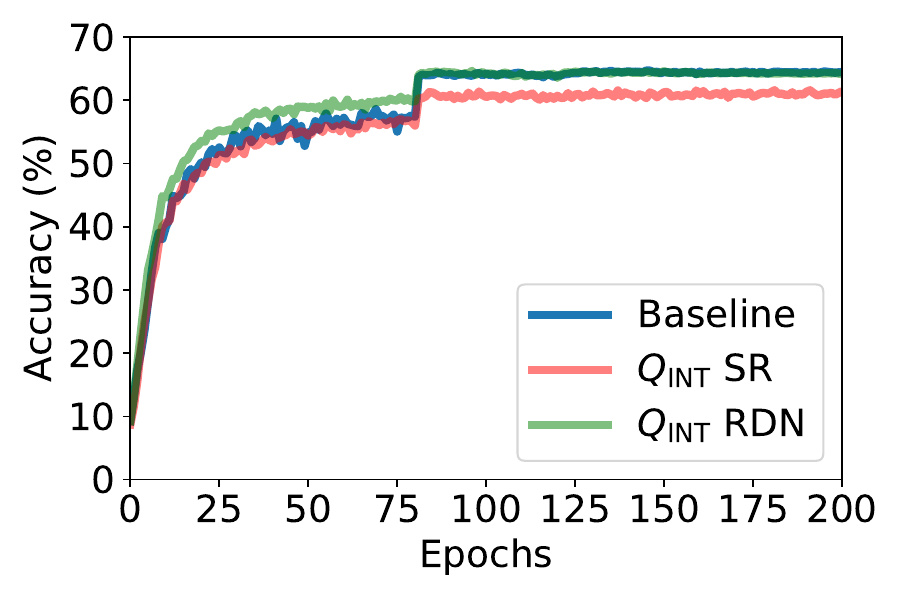}  
  \caption{}
  \label{fig:fwdSrRdn}
 \end{subfigure}
\begin{subfigure}{0.33\textwidth}
  \centering
  \includegraphics[width=\linewidth]{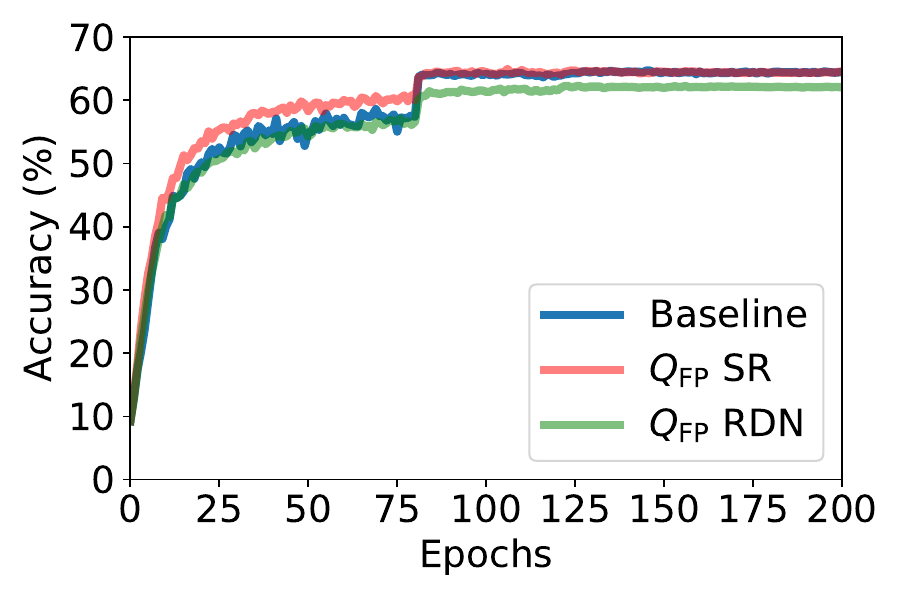}  
  \caption{}
  \label{fig:bwdSrRdn}
\end{subfigure}
\caption{Comparison between stochastic rounding (SR) and round-to-nearest (RDN) quantization. In \textbf{(a)} we present the MSE of a uniform distributed tensor with the two different rounding schemes. Quantization to 4 bits of the activations and weights ($Q_{\mathrm{INT}})$ \textbf{(b)} and neural gradients ($Q_{\mathrm{FP}}$) \textbf{(c)} of ResNet18 - Cifar100 dataset with SR and RDN. While MSE is important in $Q_{\mathrm{INT}}$ for the weights and activations, unbiasedness achieved with SR is crucial for the neural gradients in $Q_{\mathrm{FP}}$. The neural gradients in (b) and the weights and activations in (c), are in full precision to focus on the effect of the rounding scheme only in one part of the network in each experiment. }
\label{fig:FwdBwdSR}
\end{figure*}

As we explained in section \ref{sub: unbiased gradients}, unbiased neural gradients quantization leads to an unbiased estimate of the weight gradients, which enables proper convergence of SGD \citep{bottou2018optimization}. Thus, bias in the gradients can hurt the performance and should be avoided, even at the cost of increasing the MSE. Therefore, neural gradient quantization ($Q_{\mathrm{FP}}$), should be done using a SR rounding scheme, following subsection \ref{sub: unbiased gradients}.
In \cref{fig:fwdSrRdn,fig:bwdSrRdn} we see that these theoretical observations are consistent with empirical observations favoring RDN for the weights and activations ($Q_{\mathrm{INT}}$) and SR for the neural gradients ($Q_{\mathrm{FP}}$). 


\begin{figure}[h]
\begin{subfigure}[b]{0.4\linewidth}
  \centering
  \includegraphics[width=\linewidth]{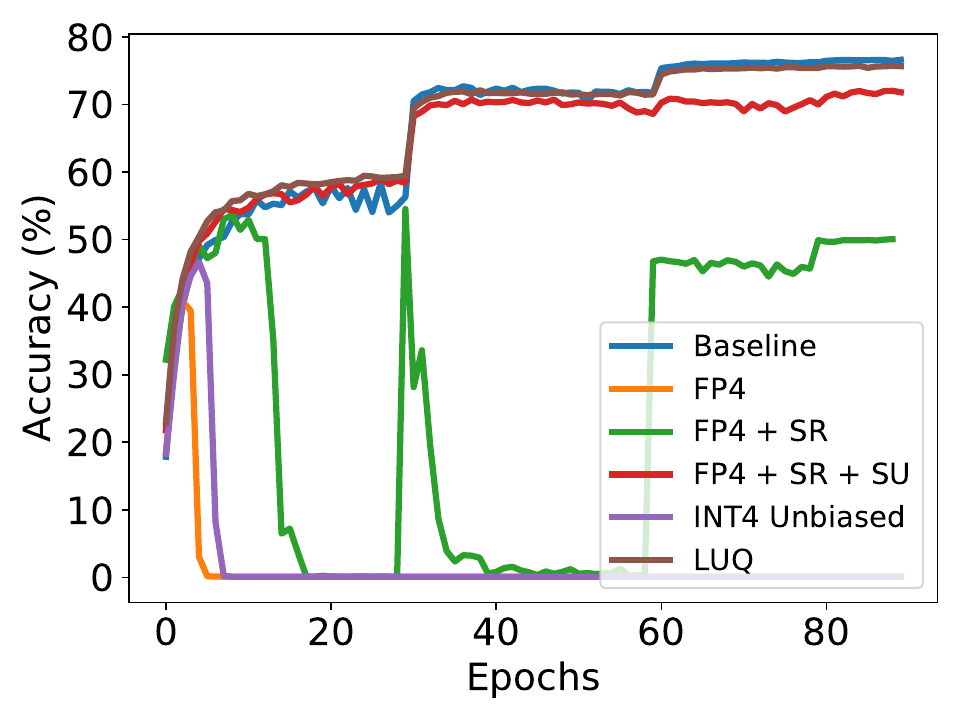}  
  \end{subfigure}
  \hfill
\begin{subfigure}[b]{0.59\linewidth}  \centering
  \includegraphics[width=\linewidth, ]{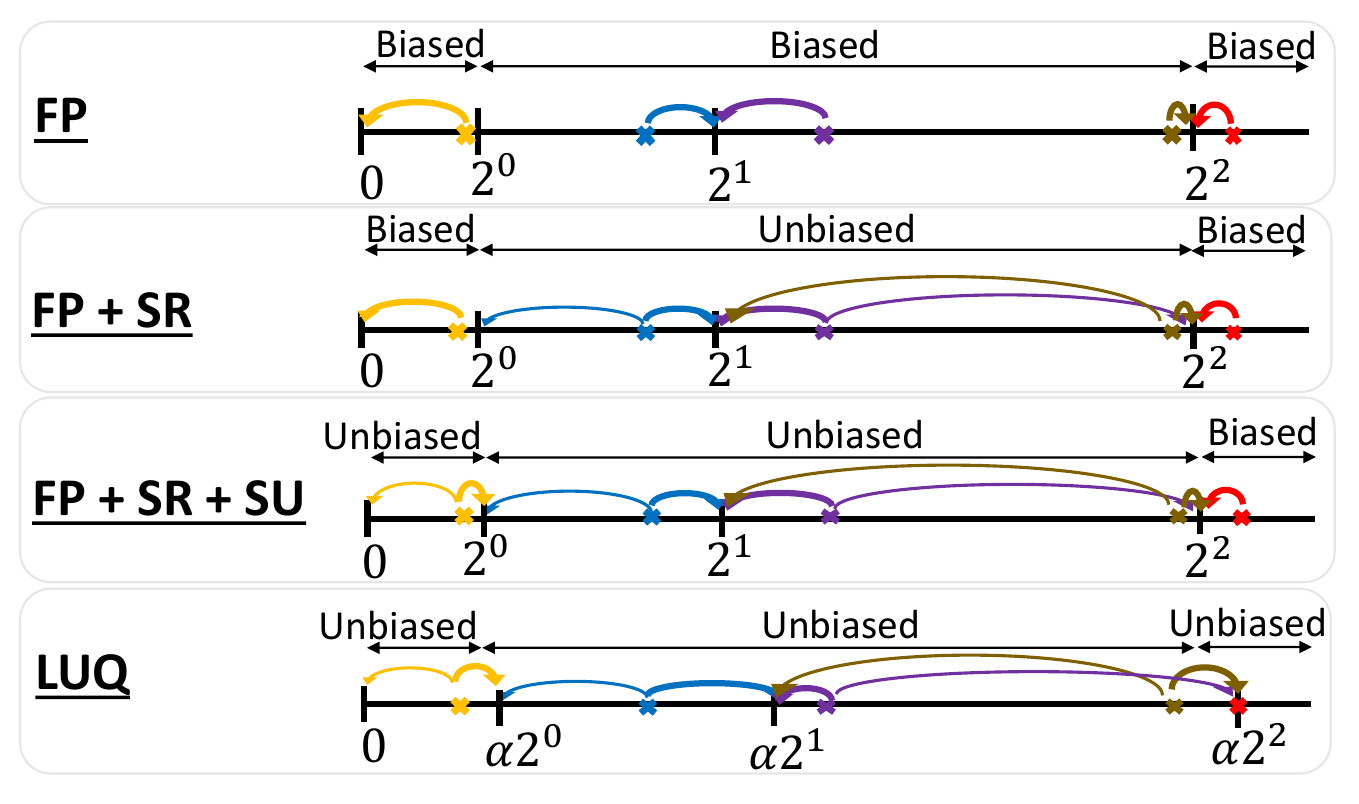}  
  \end{subfigure}
  \caption{\textbf{(Left):} ResNet50 top-1 validation accuracy in ImageNet dataset with different quantization schemes for the neural gradients. FP4 refers to standard logarithmic (1-3-0) floating point quantization. SR refers to stochastic rounding, which makes the quantization unbiased inside the FP range. SU refers to stochastic underflow which makes the quantization unbiased below minimum FP. "INT4 unbiased" refers to the combination of INT4 and SR which is fully unbiased. Notice that while biased logarithmic quantization ("FP4"), partially biased logarithmic ("FP4 + SR", "FP4 + SR + SU"), and uniform unbiased ("INT4 unbiased") lead to significant accuracy degradation, the proposed fully unbiased logarithmic quantization ("LUQ") has a minimal degradation.  \textbf{(Right):} Illustration of the different logarithmic quantization schemes for FP2 ([1,1,0] format), where $2^0$ and $2^2$ are, respectively, the minimal and maximal FP representations. Two arrows for the same point mean SR - thicker lines represent higher probability. Only LUQ is able to achieve unbiasedness in all floating point ranges.\label{fig:repetitionsAndAblations}
}
\vspace{-.6cm}
\end{figure}
\vspace{-.6cm}
\section{LUQ: a logarithmic unbiased quantizer}
\label{sec:luq}

Following the conclusions from the previous section, the neural gradients quantizer ($Q_{\mathrm{FP}}$ in eqs. \ref{eq:backward} and \ref{eq:update} ) should be logarithmic and be completely unbiased. In this section we aim to do this efficiently and create, for the first time, a Logarithmic Unbiased Quantizer (LUQ).

Standard radix-2 floating-point defines a dynamic range. In standard FP, all the values below the minimum FP representation are pruned to 0 and all the values above the maximum FP representation are clipped to the maximum. In order to create a fully unbiased quantizer, we need to keep all the following three regions unbiased: below range minimum, inside range, and above range maximum.

\paragraph{1) Below FP minimum: Stochastic underflow} 
Given an underflow threshold $\alpha$ we define a stochastic pruning operator, which prunes a given value $x$, as
\begin{equation}
 T_{\alpha}\left (x\right)=\begin{cases}
x &\quad \mathrm{, if}$  $|x|\geq\alpha\\
\text{sign}(x) \cdot \alpha & \quad w.p.  \,\,\,\frac{|x|}{\alpha},  \mathrm{ if}$  $|x|<\alpha \\
0 & \quad w.p. \,\,\, 1-\frac{|x|}{\alpha},  \mathrm{ if} $  $ |x|<\alpha \, .
\end{cases}
\label{StochasticPruning}
\end{equation}
\paragraph{2) Above FP maximum: Underflow threshold} In order to create an unbiased quantizer, the largest quantization value $2^{2^{b-1}}\alpha$ should avoid clipping any values in a tensor $x$, otherwise this will create a bias. Therefore, the maximal quantization value is chosen as $\max(|x|)$, the minimal value which will avoid clipping and bias. Accordingly, the underflow threshold $\alpha$ is (with $b=3$ for FP4)  
\begin{equation} 
\label{eq:alpha}
\alpha = \frac{\text{max}(|x|)}{2^{2^{b-1}}}\, .
\end{equation}

\rebuttal{The operation in \cref{eq:alpha} can have some overhead, but in \cref{sec:discussion} we present two methods ("Reducing the data movement" and "Reducing the cost of the scaling operation") to decrease this overhead}.
\vspace{-.2cm}
\paragraph{3) Inside FP range:  Logarithmic SR} Given an underflow threshold $\alpha$, let  $Q_{\alpha}(x)$ be a FP round-to-nearest $b$-bits quantizer with bins $\{\alpha,2\alpha,...,2^{2^{b-1}}\alpha\}$.  Assume, without loss of generality, $2^{n-1}\alpha < x < 2^n\alpha$  $( n \in \{0,1 ... ,  b-1\})$ .  We will use the following quantizer, which is a special case of SR (\cref{SR_eq}), and is unbiased as a special case of \cref{expecration_SR}.:
\begin{equation}
Q_{\alpha}(x) = \begin{cases}
2^{n-1}\alpha  & w.p. \,\,\, \frac{2^n\alpha-x}{2^n\alpha-2^{n-1}\alpha} \\
2^n\alpha  & w.p. \,\,\, 1 - \frac{2^n\alpha-x}{2^n\alpha-2^{n-1}\alpha} = \frac{x-2^{n-1}\alpha}{2^{n-1}\alpha}\,.
\end{cases}
\label{eq:unbiasedFpQ}
\end{equation}
The naive implementation of stochastic rounding can be expensive since it cannot use the standard quantizer. 
Traditionally, in order to use the standard quantizer the implementation includes simply adding uniform random noise $\epsilon\sim U[-\frac{2^{n-1}\alpha}{2},\frac{2^{n-1}\alpha}{2}]$   to $x$ and then using a round-to-nearest operation. The overhead of such stochastic rounding is typically negligible (\cref{sec:AppSRover}) in comparison to other operations in neural networks training. Moreover, it is possible to reduce any such overhead with the re-use of the random samples (\cref{sec:AppSRamort}). In our case, in order to implement a logarithmic round-to-nearest, we need to correct  an inherent bias since  $\alpha \cdot 2^{\lfloor{ \log \left(  \frac{|x|}{\alpha} \right) \rceil}} \neq \alpha \cdot \lfloor 2^{\log \left(  \frac{|x|}{\alpha} \right)}\rceil $.

For a bin $[2^{n-1}, 2^n]$, the midpoint $x_m$ is 
\begin{equation}
    x_{m} = \frac{2^n + 2^{n-1}}{2}=\frac{3}{4}\cdot 2^{n-1} \, .
\end{equation}

Therefore, we can apply round-to-nearest-power (RDNP) directly on the exponent $x$ of any value $2^{n-1}\leq 2^x \leq 2^n$ as follows:  
\vspace{-0.2cm}
\begin{equation}
\begin{split}
    \text{RDNP}(2^x) &= 2^{ \lfloor\log{\left(\frac{4}{3}\cdot 2^x\right)\rfloor}}= 2^{ \lfloor x+ \log{\left(\frac{4}{3}\right)}\rfloor} =
    2^{ \text{RDN} \left(x+ \log{\left(\frac{4}{3}\right)}-\frac{1}{2}\right)  } \approx 2^{\text{RDN} \left(x - 0.084\right)}\, .
    \label{eq:RC}
\end{split}
\end{equation}

\rebuttal{Notice that the use of RDNP avoids the need of converting back to linear space in order to implement SR and avoid additional computational overhead.}


\paragraph{Logarithmic unbiased quantization (LUQ)}

LUQ, the quantization method we suggested above, is unbiased since it can be thought of as applying logarithmic stochastic rounding (\cref{eq:unbiasedFpQ}) on top of stochastic pruning (\cref{StochasticPruning}) 
\begin{equation}
    X_q = Q_{\alpha}\left (T_{\alpha}(x)\right)\, .
\end{equation}
Since $T_{\alpha}$ and $Q_{\alpha}$ are unbiased, $X_q$ is an unbiased estimator for $x$, from the law of total expectation,
\begin{equation}
\begin{split}
E[X_q] &= E\left[ Q_{\alpha}\left (T_{\alpha}(x)\right)\right] =  E\left[ E\left[ Q_{\alpha}\left (T_{\alpha}(x)\right)\right]|T_{\alpha}(x) \right] = E\left [T_{\alpha}(x) \right]=x\, ,
\end{split}
\end{equation}
where the expectation is over the randomness of $T_{\alpha}$ and $Q_{\alpha}$. 

In \cref{fig:repetitionsAndAblations} (Left) we show an ablation study of the effect of the different quantization schemes on ResNet50 in ImageNet: while standard (biased) FP4 diverges, adding stochastic rounding or stochastic underflow (which make the process partially unbiased) enables convergence, but with significant degradation. Combining logarithmic quantization with full unbiasedness in LUQ obtained minimal accuracy degradation. Notice also that only unbiasedness without logarithmic quantization ("INT4 unbias") completely diverges. In \cref{fig:repetitionsAndAblations} (Right) we show an illustration of the different logarithmic quantization schemes, where only LUQ achieved a fully logarithmic unbiasedness FP quantization. 
\vspace{-0.2cm}
\section{Optional methods} \label{sec:optional methods}
\vspace{-0.1cm}
Next, we present two optional methods to improve accuracy at some computational cost.
\vspace{-0.1cm}
\subsection{SMP: Reducing the variance while keeping it unbiased}
\label{sec:rpt}

 In the previous section, we presented an unbiased method for logarithmic quantization of the neural gradients called LUQ. Following the bias-variance decomposition, if the gradients are now unbiased, then the only remaining issue should be their variance. Therefore, we suggest an optional method to reduce the quantization variance by repeatedly sampling from the stochastic quantizers in LUQ, and averaging the resulting samples of the final weight gradients. The proposed sampling can be implemented serially or in parallel. The serial implementation (re-using the same quantizer) has a power and throughput overhead but does not requires additional hardware (area) support, so it should be used if the chip area is the bottleneck. The parallel implementation avoids almost completely the throughput overhead (except the averaging operation), but it requires additional area for the multiple quantizers, so it should be used when the throughput is the bottleneck. For $N$ different samples, the proposed method will reduce the variance by a factor of $\frac{1}{N}$, without affecting the bias \citep{GILLI2019103}. In Appendix \cref{fig:repetitions} we show the effect of the different number of samples (SMP) on 2-bit quantization of ResNet18 Cifar100 dataset. There, with 16 samples, we achieve accuracy similar to a full-precision network. This demonstrates that the variance is the only remaining issue in neural gradient quantization using LUQ and that the proposed averaging method can erase this variance gap, with some overhead.  
 
 A different approach to reduce the variance can be to increase the bitwidth in the update phase. In order to keep using standard formats, we should increase the update phase to FP8-INT4, which leads to $3.5 \times$ degradation in compute density in comparison to FP4-INT4 as shown in \cite{UltraLowP4} (Table s-1). Therefore, using the proposed SMP with two samples (as we shall do) has a significant advantage in compute density ($\1.75 \times$) in comparison to increasing the bitwidth.

\subsection{FNT: Fine-tuning in high precision}
\label{sec:fnt}
After the 4-bit training is finished, we suggest an optional method to reduce the gap from the full precision model, by running $T$ additional iterations in which we increase all the network parts to higher precision, except the weights which remain in low precision. We noticed that with this scheme we get the best accuracy for the fine-tuned network. At inference time the activations and weights are quantized to lower precision. During the fine-tune phase, the Learning Rate (LR) is increased linearly during $\frac{T}{2}$ iterations and then reduced linearly with the same slope:
 \begin{equation}
 \text{LR}_t =\begin{cases}
\text{LR}_{T} + \frac{(\text{LR}_{\text{base}} - \text{LR}_{T})}{T/2} t &\quad \mathrm{, if}$  $t \leq \frac{T}{2}  \\
\text{LR}_\text{base} +  \frac{(\text{LR}_{T}-\text{LR}_\text{base})}{T/2}  t  &\quad  \mathrm{, else} \, 
\end{cases}\, ,
\label{StochasticPruning}
\end{equation}
where $\text{LR}_T$ is the final LR of the 4-bit training and $\text{LR}_{\mathrm{base}}$ is the maximal LR of the fine-tune phase. 

\section{Experiments}

In this section, we evaluate the proposed LUQ for 4-bit training on various DNN models. For all models, we use their default architecture, hyper-parameters, and optimizers combined with a custom-modified Pytorch framework that implemented all the low precision schemes. Additional experimental details appear in \cref{sec:appExp}. 
\vspace{-0.3cm}

\paragraph{INT4 quantization}
INT4 quantization methods for the weights and activations (forward pass) were well studied in the past. In this paper, we used SAWB \cite{SAWB} to quantize the weights and activations. SAWB determines the quantization scaling factor by first finding the optimal (in terms of MSE) scaling factor on six distribution approximations of the true tensor distribution, and then applying linear regression to find the chosen scaling factor.  

\paragraph{Training time measurement} Notice that, currently, AI accelerators do not support 4-bit formats for training. This means that we can only simulate the quantization process, but are not able to measure training time or memory reduction. This is the common practice in the neural network quantization literature, where the algorithms often appear before the hardware that can support them. For example, though we can find FP8 training publications since 2019 \citep{Sun2019Hybrid8F},  only recently did Nvidia announce their first GPU that supports the FP8 format (H100).

\paragraph{Main results} In \cref{tab:exp} we show the Top-1 accuracy achieved in 4-bit training using LUQ to quantize the neural gradients to FP4 and combined with a previously suggested method, SAWB \citep{SAWB},  to quantize the weights and activations to INT4. We compare our method with Ultra-low \citep{UltraLowP4} showing better results in all the models, achieving SOTA in 4-bit training. Moreover, we improve the results by using the proposed SMP (\cref{sec:rpt}). In \cref{tab:expFNT} we show the effect of the proposed fine-tuning, reducing or closing completely the gap from full-precision model. We verified that stochasticity has only a negligible effect on the variance of final performance by running a few different seeds. Additional experiments appear in \cref{sec:AppAdditionalexp}.

\begin{table*}[h!]
\centering
\caption{Comparison of 4-bit training of the proposed method LUQ with Ultra-low \citep{UltraLowP4} in various vision models with ImageNet dataset, Transformer-base in WMT En-De task dataset and BERT fine-tune in SQUAD dataset. SMP refers to doing two samples of the SR quantization of neural gradients in order to reduce the variance (\cref{sec:rpt}).}
\label{tab:exp}
\begin{tabular}{lcccc}
\toprule
\multicolumn{1}{l|}{Model} & \multicolumn{1}{c|}{Baseline} & \multicolumn{1}{c|}{Ultra-low \footnotemark \citep{UltraLowP4} } & \multicolumn{1}{c|}{LUQ} &  \multicolumn{1}{c}{LUQ + SMP}  \\ \midrule
\multicolumn{1}{l|}{ResNet-18} & \multicolumn{1}{c|}{69.7 \%} & \multicolumn{1}{c|}{68.27\%} & \multicolumn{1}{c|}{69.09\%} &  \multicolumn{1}{c}{69.24 \%}\\ \hline
\multicolumn{1}{l|}{ResNet-50} & \multicolumn{1}{c|}{76.5\%} & \multicolumn{1}{c|}{74.01\%} & \multicolumn{1}{l|}{75.42 \%} &    \multicolumn{1}{c}{75.63 \%} \\ \hline
\multicolumn{1}{l|}{MobileNet-V2} & \multicolumn{1}{c|}{71.9 \%} & \multicolumn{1}{c|}{68.85 \%} & \multicolumn{1}{c|}{69.55 \%} &  \multicolumn{1}{c}{69.7 \%}  \\ \hline
\multicolumn{1}{l|}{ResNext-50} & \multicolumn{1}{c|}{77.6 \%} & \multicolumn{1}{c|}{N/A} & \multicolumn{1}{c|}{76.02 \%} & \multicolumn{1}{c}{76.12 \%} \\ \hline
\multicolumn{1}{l|}{Transfomer-base} & \multicolumn{1}{c|}{27.5 (BLEU)} & \multicolumn{1}{c|}{25.4} & \multicolumn{1}{c|}{27.17} & \multicolumn{1}{c}{27.25} \\ \hline
\multicolumn{1}{l|}{BERT fine-tune} & \multicolumn{1}{c|}{87.03 (F1)} & \multicolumn{1}{c|}{N/A} & \multicolumn{1}{c|}{85.75} & \multicolumn{1}{c}{85.9} \\   \hline
\multicolumn{1}{l|}{ViT B} & \multicolumn{1}{c|}{76.2} & \multicolumn{1}{c|}{N/A} & \multicolumn{1}{c|}{73.7 \%} & \multicolumn{1}{c}{74.1 \%} \\ 
\bottomrule
\end{tabular}

\end{table*}
\footnotetext[3]{Recall Ultra-low used a non-standard radix-4 quantization format, that significantly reduces the benefit of using low bit quantization. }
\vspace{-0.3cm}

\begin{table*}[h!]
\centering
\caption{Effect of the proposed FNT method (\cref{sec:fnt}) using FP16 format with different epochs.} 
\label{tab:expFNT}
\begin{tabular}{lccccc}
\toprule
\multicolumn{1}{l|}{Model} & \multicolumn{1}{c|}{Baseline} & \multicolumn{1}{c|}{LUQ + SMP} & \multicolumn{1}{c|}{+FNT 1 epoch} &  \multicolumn{1}{c|}{+FNT 2 epochs} &  \multicolumn{1}{c}{+FNT 3 epochs} \\ \midrule
\multicolumn{1}{l|}{ResNet-18} & \multicolumn{1}{c|}{69.7 \%} & \multicolumn{1}{c|}{69.24 \%} & \multicolumn{1}{c|}{69.7 \%} &  \multicolumn{1}{c|}{-} & 
\multicolumn{1}{c}{-} \\ \hline
\multicolumn{1}{l|}{ResNet-50} & \multicolumn{1}{c|}{76.5 \%} & \multicolumn{1}{c|}{75. 63\%} & \multicolumn{1}{c|}{75.89 \%} &  \multicolumn{1}{c|}{76 \%} & 
\multicolumn{1}{c}{76.18 \%} \\ \hline
\multicolumn{1}{l|}{MobileNet-V2} & \multicolumn{1}{c|}{71.9 \%} & \multicolumn{1}{c|}{69.7 \%} & \multicolumn{1}{c|}{70.1 \%} &  \multicolumn{1}{c|}{70.3 \%} & 
\multicolumn{1}{c}{70.3 \%} \\ \hline
\multicolumn{1}{l|}{ResNext-50} & \multicolumn{1}{c|}{77.6 \%} & \multicolumn{1}{c|}{76.12\%} & \multicolumn{1}{c|}{76.25 \%} &  \multicolumn{1}{c|}{76.33 \%} & 
\multicolumn{1}{c}{76.7 \%} \\
\bottomrule
\end{tabular}
\end{table*}

\paragraph{Overhead of SMP and FNT} We limit our experiments with the proposed SMP method to only two samples. This is to achieve a similar computational overhead as Ultra-low\citep{UltraLowP4}, with their suggested two-phase-rounding (TPR) which also generates a duplication for the neural gradient quantization.  Additional ablation study of the SMP overhead appears in \cref{sec:AppAdditionalexp}. The throughput of a 4-bit training network is approximately 8x in comparison to full precision training  \citep{UltraLowP4}. This means that doing one additional epoch in high precision reduces the throughput by $\sim 8 \%$. In comparison, Ultra-low \citep{UltraLowP4} does full-training with all the 1x1 convolutions in 8bit, which reduces the throughput by $\sim 50 \%$ in comparison to all 4bit training. 
\paragraph{Forward-backward ablations} In Appendix \cref{tab:FwdvsBwd} we show the Top-1 accuracy in ResNet50 with different quantization schemes. The forward phase (activations + weights) is quantized to INT4 with SAWB \citep{SAWB} and the backward phase (neural gradients) to FP4 with LUQ. As expected, the network is more sensitive to the quantization of the backward phase.

\vspace{-0.1cm}
\section{Discussion} \label{sec:discussion}
\vspace{-0.1cm}

\paragraph{Conclusions}

In this work, we analyze the difference between two rounding schemes: round-to-nearest and stochastic-rounding. We showed that, while the former has lower MSE and works better for the quantization of the forward phase (weights and activations), the latter is an unbiased approximation of the original data and works better for the quantization of the backward phase (specifically, the neural gradients). 

Based on these conclusions and previous works (\cite{Chmiel2020NeuralGA}) that showed the optimally of logarithmic quantization, we propose the first method that combined logarithmic quantization with unbiasedness, with the proposed logarithmic unbiased quantizer (LUQ) which quantize the neural gradients to format FP4 [1,3,0].  Combined with a known method for quantizing the weights and activations to INT4 we achieved, without overhead, state-of-the-art in standard format 4-bit training in all the models we examined, e.g., 1.1\% error in ResNet50 vs. 2.49\% for the previous known SOTA (\cite{UltraLowP4}, which used non-standard format). 

Moreover, we suggest two more methods to improve the results, with overhead comparable to \cite{UltraLowP4}. The first reduces the quantization variance, without affecting the unbiasedness of LUQ, by averaging several samples of stochastic neural gradients quantization. The second is a simple method for fine-tuning in high precision for one epoch. Combining all these methods, we were able for the first time to achieve 0.32\% error in 4-bit training of ResNet50 ImageNet dataset. 
\vspace{-0.3cm}

\paragraph{Reducing the data movement}
So far, we focused on resolving the 4-bit GEMM operation bottleneck in DNNs training. It reduces not only the computational resources for the GEMM operations, but also reduces the required memory in DNNs training.  However, LUQ, similarly to previous quantization methods \citep{UltraLowP4,PACT,SAWB}, requires a statistical measurement of the tensor to define the quantization dynamic range. Specifically, LUQ requires a measurement of the maximum of the neural gradient tensor. Such measurement increases the data movement from and to memory, making this data movement a potential bottleneck in some hardware. 

In order to avoid this bottleneck when it is a major issue, we verified that LUQ can be combined with the In-hindsight \cite{Fournarakis2021InHindsightQR} statistics estimation method, which uses a pre-computed measurement to quantize the current tensor and in parallel extract the current statistics for the next iteration. The maximum estimate in LUQ, $\hat{m}$ is calculated as:
\begin{equation}
    \hat{m}^t = (1-\eta)\cdot\text{max}(|x^{t-1}|) + \eta \cdot \hat{m}^{t-1}\, ,
\end{equation}
where $\eta$ is the momentum hyperparameter, and $x$ is tensor of neural gradients. In \cref{sec:AppHindsight} we show the effectiveness of this statistic estimation, which eliminates this data movement bottleneck, with a negligible change to the accuracy. Though this method can potentially introduce some bias to the quantization process, the bias seems to be negligible (see Appendix \cref{fig:hinsightMxMeasure}).
\vspace{-0.3cm}
\paragraph{Reducing the cost of the scaling operation} 
In \cref{sec:appHWFriend} we present an optional method to convert the underflow threshold $\alpha$ (\cref{eq:alpha}) to a power-of-two, at a small cost to accuracy. This can reduce or eliminate the computational overhead of the multiplication by $\alpha$ in eq. \cref{eq:unbiasedFpQ}, which can be significant in some cases.
\vspace{-0.3cm}
\paragraph{Multiplication free backpropagation} 
In this work, we reduce the GEMM bottleneck by combining two different data-types for the forward (INT4) and backward (FP4) passes. Standard GEMM operations with different formats, require casting the operand to a common datatype before the multiplication. The cost of the casting operation can be significant. We notice, that we are dealing with a special case, where one of the operands includes only a mantissa (weights and activations) and the other only an exponent (neural gradients). Our initial analysis (\cref{sec:mfBprop}) shows that this allows, with small hardware changes, reduction of the area of standard GEMM block by $5\times$. 
\vspace{-0.3cm}
\paragraph{Accumulation width} A different future direction is to reduce the accumulator width, which is usually kept as FP32. As explained in \cref{sec:mfBprop}, the FP32 accumulator is the most expensive block when training in low bits. Now, after allowing training with 4-bit it is reasonable to think that the accumulator width can be reduced. 

\newpage

\section*{Acknowledgement }

The research of DS was Funded by the European Union (ERC, A-B-C-Deep, 101039436). Views and opinions expressed are however those of the author only and do not necessarily reflect those of the European Union or the European Research Council Executive Agency (ERCEA). Neither the European Union nor the granting authority can be held responsible for them. DS also acknowledges the support of Schmidt Career Advancement Chair in AI.

\bibliography{iclr2023_conference}
\bibliographystyle{iclr2023_conference}
\newpage
\appendix
\section{Appendix}

\subsection{Quantizations formats}
\label{sec:AppQFormats}
\begin{figure}[h]
  \centering
  \includegraphics[width=1\linewidth]{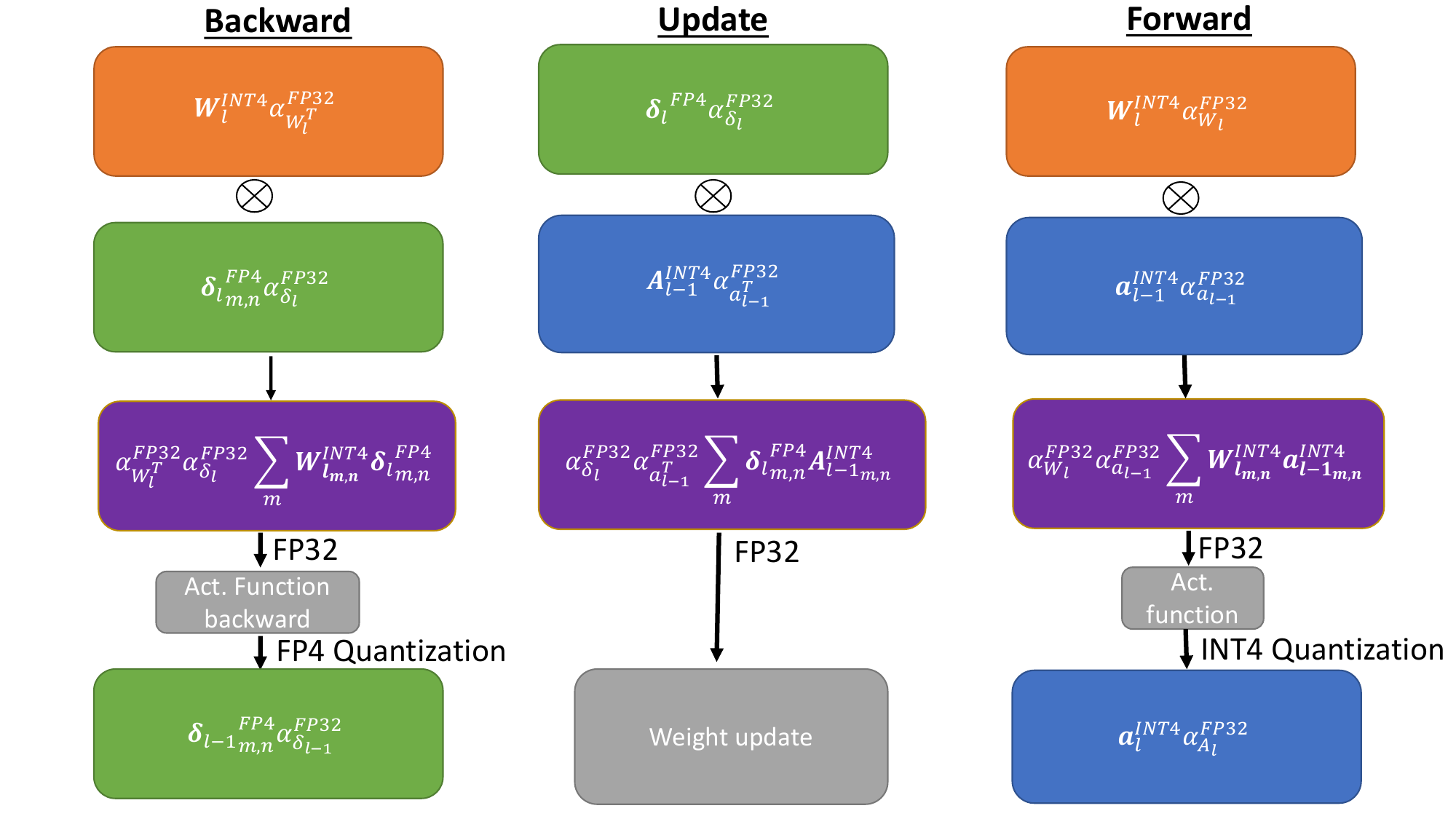}  
  \caption{Summary of the different formats used in each of the three GEMM in our 4 training regime. Boldface represent matrix and vectors. Different colors represent different elements of each layer (activations, weights, neural gradients, etc.)}
\label{fig:quantizeFormats}
\end{figure}

In this paper, we aim to reduce the General matrix multiplication (GEMM) to 4 bit. The activation and weights are quantized to standard INT4, while the neural gradients to FP4 with LUQ. In the forward path (\cref{eq:forward}), we use standard INT quantization (Equation 3 in \cite{Nagel2021AWP}):
$$z_l = \sum Q_{INT4}(W_l)Q_{INT4}(a_l) = \sum \alpha_{W_l}W_l^{INT4}\alpha_{a_l}a_l^{INT4} = \alpha_{W_l}\alpha_{a_l}\sum W_l^{INT4}a_l^{INT4} \, ,$$ where $\alpha$ is the INT scale factor and defined as:  $\alpha = \frac{\Delta}{2^b-1}\,$ where $\Delta$ is the quantized range and $b$ is the quantization bitwidth. In this standard INT quantization all MAC operations are in low precision and only the final multiplication with the scale requires high precision. For the backward (\cref{eq:backward}) and update (\cref{eq:update}) GEMMs we use exactly the same scheme: again the scale ($\alpha$) is in high precision but since it is the same for all the tensor, all MAC operation can be done in low precision. The only difference is that we require a MAC unit that performs INT4-FP4 operations. This requirement is the same as required in Ultra-low (\cite{UltraLowP4}) and as they showed, the support of such INT4-FP4 MAC unit is simple and requires 55\% of the area of standard FP16 Floating-point-unit, while providing a 4x throughput. In \cref{fig:quantizeFormats} we show all these details.

\subsection{Comparison of mean-square-error - full derivatives}
\label{sec:appFullMSE}

Given that we want to quantize $x$ in a bin with a lower limit $l(x)$ and an upper limit $u(x)$, stochastic rounding can be stated as follows:
\begin{equation}
\label{SR_eqApp}
\mathrm{SR}(x)= \begin{cases}l(x) , &  w.p. \quad p(x) = 1-\frac{x-l(x)}{u(x)-l(x)}\\ u(x), & w.p. \quad 1-p(x) = \frac{x-l(x)}{u(x)-l(x)}\end{cases} \, .
\end{equation}

The expected rounding value is given by
\begin{equation}
\label{expecration_SRApp}
\begin{split}
    E[\mathrm{SR}(x)] &= l(x) \cdot p(x) + u(x)\cdot (1-p(x)) = x\, ,
\end{split}
\end{equation}
where here and below the expectation is over the randomness of SR (i.e., $x$ is a deterministic constant).

Stochastic rounding is an unbiased approximation of $x$, since it has
zero bias: 
\begin{equation}
    \operatorname{Bias}[\mathrm{SR}(x)] = E[\mathrm{SR}(x) - x]  = E[\mathrm{SR}(x)] -x  = 0 \,.
\end{equation}
However, stochastic rounding has a variance, given by
\begin{equation}
\begin{split}
\label{VAR_SR}
\operatorname{Var}[\mathrm{SR}(x)]& = (l(x)-E[\mathrm{SR}(x)])^{2} \cdot p(x) +  (u(x)-E[\mathrm{SR}(x)])^{2}\cdot (1-p(x))  \\
&=\left(x-l(x)\right)\cdot \left(u(x)-x\right) \,,
\end{split}
\end{equation}
where the last transition follows from substituting the terms $E[\mathrm{SR}(x)])$, and $p(x)$ into \cref{VAR_SR}.

We turn to consider the round-to-nearest method (RDN). The bias of RDN is given by
\begin{equation}
\operatorname{Bias}[\mathrm{RDN}(x)] = \min{\left(x - l(x) , u(x) -  x \right)}\, .
\end{equation}
Since RDN is a deterministic method, it is evident that the variance is 0 i.e., 
\begin{equation}
\operatorname{Var}[\mathrm{RDN}(x)]= 0\,.
\end{equation}
Finally, for every value $x$ and a rounding method $R(x)$, the mean-square-error (MSE) can be written as the sum of the rounding variance and the squared rounding bias,
\begin{equation}
    \text{MSE}[R(x)] = E[R(x) - x]^2 = \operatorname{Var}[\mathrm{R}(x)] +  \operatorname{Bias}^2[\mathrm{R}(x)]\, .
\end{equation}
Therefore, we have the following MSE distortion when using round-to-nearest and stochastic rounding:
\begin{equation}
 \text{MSE}=\begin{cases}
\left[\min{\left(x - l(x) , u(x) -  x \right)}\right]^2 &\quad RDN(x)\\
\left(x-l(x)\right)\cdot \left(u(x)-x\right) & \quad SR(x) \\
\end{cases} \, .
\label{MSE_}
\end{equation}

\subsection{When is it important to use unbiased quantization - full derivatives}
\label{sec:appFullUnbias}

Denote $W_l$ as the weights between layer $l-1$ and $l$, $C$ as the cost function, and $f_l$ as the activation function at layer $l$. Given an input–output pair $(x,y)$, the loss is:
\begin{equation}
\label{fwd}
C\left(y, f_{L}\left(W_{L} f_{L-1}\left(W_{L-1} \cdots f_{2}\left(W_{2} f_{1}\left(W_{1} x\right)\right) \cdots\right)\right)\right) \, .
\end{equation}

\textbf{Backward pass.}  Let $z^{l}$ be the weighted input (pre-activation) of layer $l$ and denote the output (activation) of layer $l$ by $a_{l}$. The derivative of the loss in terms of the inputs is given by the chain rule:

\begin{equation}
\label{Backprop}
\delta_{l}\triangleq\frac{dC}{dz_{l}}=\frac{da_{l}}{dz_{l}}\cdots\frac{dz_{L-1}}{da_{L-2}}\cdot\frac{da_{L-1}}{dz_{L-1}}\cdot\frac{dz_{L}}{da_{L-1}}\cdot\frac{da_{L}}{dz_{L}} \cdot \frac{dC}{da_{L}} \, .
\end{equation}
Therefore, $\delta_{L}\triangleq\frac{da_{L}}{dz_{L}}\cdot \frac{dC}{da_{L}}$, and we can write recursively the backprop rule $\forall l<L$:
\begin{equation}
\label{Backprop}
\delta_{l}\triangleq\frac{da_{l}}{dz_{l}}\frac{dz_{l+1}}{da_{l}}\cdot\delta_{l+1} \, .
\end{equation}
In its quantized version, $\delta_{L}^{q}=Q\left(\delta_{L}\right)$, and \cref{Backprop} has the following form (with ReLU activations): 

\begin{equation}
\label{quantizedBackprop}
\delta_{l}^{q}\triangleq Q\left(\frac{da_{l}}{dz_{l}}\frac{dz_{l+1}}{da_{l}}\delta_{l+1}^{q}\right) \, .
\end{equation}
Assuming $Q(x)$ is an unbiased stochastic quantizer with $E[Q(x)]=x$, we next show the quantized backprop $\delta_{l}^q$ is an unbiased approximation of backprop: 

\begin{equation}
\begin{split}    
\label{quantizedBWD}
E\delta_{l}^{q} &= EQ\left(\frac{da_{l}}{dz_{l}}\frac{dz_{l+1}}{da_{l}}\delta_{l+1}^{q}\right)  \overset{\left(1\right)}{=}   E\left[E\left[Q\left(\frac{da_{l}}{dz_{l}}\frac{dz_{l+1}}{da_{l}}\delta_{l+1}^{q}\right)|\delta_{l+1}^{q}\right]\right]\\
 &\overset{\left(2\right)}{=}E\left[\frac{da_{l}}{dz_{l}}\frac{dz_{l+1}}{da_{l}}\delta_{l+1}^{q}\right]\overset{\left(3\right)}{=}\frac{da_{l}}{dz_{l}}\frac{dz_{l+1}}{da_{l}}E\delta_{l+1}^{q} \overset{\left(4\right)}{=}\frac{da_{l}}{dz_{l}}\frac{dz_{l+1}}{da_{l}}\delta_{l}
 \, ,
 \end{split}    
\end{equation}
where in (1) we used the law of total expectation, in (2) we used $E[Q(x)]=x$, in (3) we used the \textit{linearity} of back-propagation, and in (4) we assumed by induction that $E\left[\delta_{l+1}\right]=\delta_{l+1}^{q}$, which holds initially: $E\delta_{L}^{q}=EQ\left({\delta_{L}}\right)=\delta_{L}$.

Finally, the gradient of the weights in layer $l$ is $\nabla_{W_{l}} C=\delta_{l}a_{l-1}^{\top}$ and its quantized form is $\nabla_{W_{l}} C_q=\delta_{l}^q  a_{l-1}^{\top}$. Therefore, the update $\nabla_{W_{l}} C_q$ is an unbiased estimator of $\nabla_{W_{l}} C$:
\begin{equation}
\begin{split}
\label{update}
    E\left[\nabla_{W_{l}} C_q\right]&=E\left[\delta_{l}^q  a_{l-1}^{\top}\right] = E\left[\delta_{l}^q\right]a_{l-1}^{\top} = \delta_{l} a_{l-1}^{\top} =  \nabla_{W_{l}} C \, .
\end{split}
\end{equation}

\subsection{Experimental details}
\label{sec:appExp}

In all our experiments we use the most common approach \citep{Banner2018ScalableMF,PACT} for quantization where a high precision of the weights is kept and quantized on-the-fly. The updates are done in full precision. In all our experiments we use 8 GPU GeForce Titan Xp or GeForce RTX 2080 Ti or Ampere A40.

\paragraph{ResNet / ResNext} We run the models ResNet-18, ResNet-50 and ResNext-50 from torchvision. We use the standard pre-processing of ImageNet ILSVRC2012 dataset. We train for 90 epochs, use an initial learning rate of 0.1 with a 0.1 decay at epochs 30,60,80. We use standard SGD with momentum of 0.9 and weight decay of 1e-4. The minibatch size used is 256. Following the DNNs quantization conventions \citep{Banner2018ScalableMF,Nahshan2019LossAP,PACT} we kept the first and last layer (FC) at higher precision. Additionally, similar to \cite{UltraLowP4} we adopt the full precision at the shortcut which constitutes only a small amount of the computations ($\sim 1\%$). We totally The "underflow threshold" in LUQ is updated in every bwd pass as part of the quantization of the neural gradients. In all experiments, the BN \rebuttal{and pooling} are calculated in high-precision. The hindsight momentum is $\eta = 0.1$ and in the FNT experiments we use $\text{lr}_{base} = 1e-3$.

\paragraph {MobileNet V2}
We run Mobilenet V2 model from torchvision. We use the standard pre-processing of ImageNet ILSVRC2012 dataset. We train for 150 epochs, use an initial learning rate of 0.05 with a cosine learning scheduler. We use standard SGD with momentum of 0.9 and weight decay of 4e-5. The minibatch size used is 256. Following the DNNs quantization conventions \citep{Banner2018ScalableMF,Nahshan2019LossAP,PACT} we kept the first and last layer (FC) at higher precision. Additionally, similar to \cite{UltraLowP4} we adopt the full precision at the depthwise layer which constitutes only a small amount of the computations ($\sim 3\%$). The "underflow threshold" in LUQ is updated in every bwd pass as part of the quantization of the neural gradients.  In all experiments, the BN \rebuttal{and pooling} are calculated in high-precision.

\paragraph {Transformer} We run the Transformer-base model based on the Fairseq implementation on the WMT 14 En-De translation task. We use the standard hyperparameters of Fairseq including Adam optimizer. We implement LUQ over all attention and feed forward layers.

\subsection{Additional experiments}
 \label{sec:AppAdditionalexp}
 
  \subsubsection{Stochastic rounding overhead}
  \label{sec:AppSRover}
 
 Efficient HW implementation of SR can be found in many accelerators that were built specifically for deep learning (Habana-Intel \cite{Habana}, Graphcore \cite{Graphcore}, Tesla \cite{Tesla}).
 In order to show the throughput overhead of stochastic rounding, when it is not supported in HW as in GPU, we measured the time of our stochastic quantizer in comparison to a deterministic quantizer (round-to-nearest) for different sizes of random tensor. The quantizer was implemented as a CUDA kernel and the experiment run on 1 A40 GPU. For each tensor size, we repeat the experiment 100 times and average the results. We see that the implementation of SR in software with our non-optimized kernel has a small overhead. This can be an option when it is not supported in HW. 

\begin{table*}[h!]
\centering
\begin{tabular}{ccc}
\toprule
\multicolumn{1}{c|}{Tensor size} & \multicolumn{1}{c|}{Stochastic rounding [micro sec] } & \multicolumn{1}{c}{Round-to-nearest [micro sec] } \\ \midrule

\multicolumn{1}{c|}{$10^3$} & \multicolumn{1}{c|}{2.64} & \multicolumn{1}{c}{2.55} \\
\hline
\multicolumn{1}{c|}{$10^4$} & \multicolumn{1}{c|}{2.64} & \multicolumn{1}{c}{2.6} \\
\hline
\multicolumn{1}{c|}{$10^5$} & \multicolumn{1}{c|}{3.94} & \multicolumn{1}{c}{3.89} \\
\hline
\multicolumn{1}{c|}{$10^6$} & \multicolumn{1}{c|}{4.07} & \multicolumn{1}{c}{4.01} \\
\hline
\multicolumn{1}{c|}{$10^7$} & \multicolumn{1}{c|}{6.89} & \multicolumn{1}{c}{6.78} \\
\hline
\multicolumn{1}{c|}{$10^8$} & \multicolumn{1}{c|}{9.83} & \multicolumn{1}{c}{9.77} \\
\bottomrule
\end{tabular}
\end{table*}

 \subsubsection{Stochastic rounding amortization}
  \label{sec:AppSRamort}
\begin{figure}[h]
  \centering
  \includegraphics[width=0.6\linewidth]{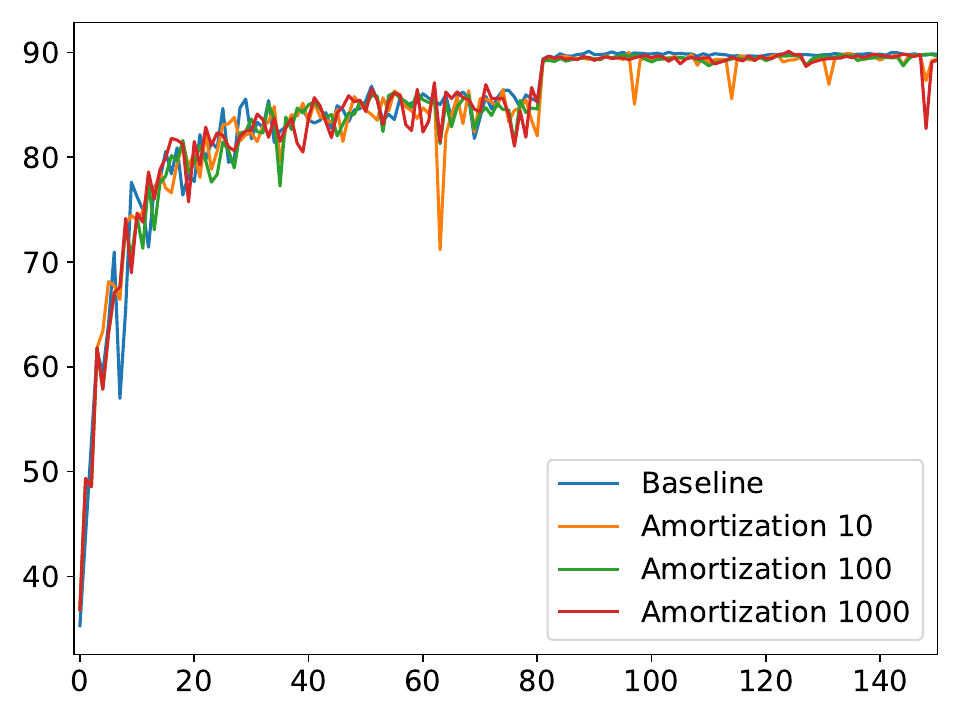}  
  \caption{ResNet18 top-1 validation in Cifar10 dataset, with 4-bit quantization of the neural gradients using stochastic-rounding. The amortization is the numbers of iteration that we re-use the random samples.}
    \label{fig:Amort}
\end{figure}

  As explained in \cref{sec:AppSRover}, usually the overhead of the stochastic rounding is typically negligible in comparison to other operations in neural network training. However, to reduce even more this overhead, is it possible to re-use the random samples. In \cref{fig:Amort} we show the effect of such re-using does not change the network accuracy.

\subsubsection {SMP experiment} 
\label{sec:appSmpExp}

\begin{figure}
\centering
  \includegraphics[width=.7\linewidth]{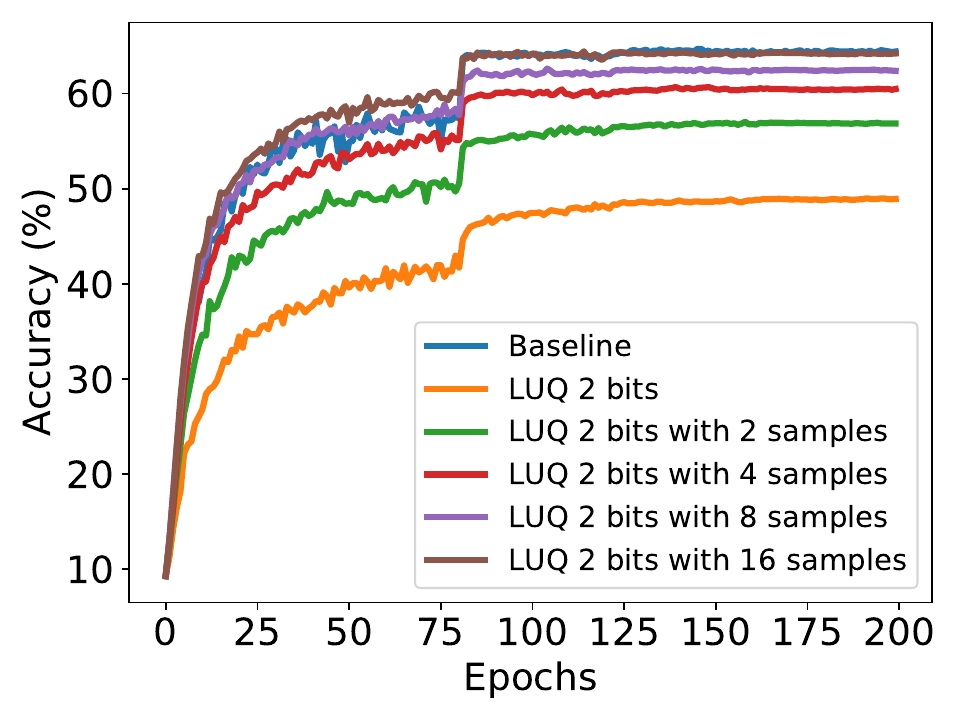}  
  \caption{ ResNet18 top-1 validation accuracy in CIFAR100 with quantization of the neural gradients to 2-bit (FP2 - [1,1,0] format) using different samples numbers to reduce the variance. Notice that 16 samples completely close the gap to the baseline. }
    \label{fig:repetitions}

\end{figure}

In \cref{fig:repetitions} we show the effect of the different number of samples (SMP) on 2-bit quantization of ResNet18 Cifar100 dataset. There, we achieve with 16 samples accuracy similar to a full-precision network. This demonstrates that the variance is the only remaining issue in neural gradient quantization using LUQ, and that the proposed averaging method can erase this variance gap, with some overhead.

\subsubsection {SMP overhead} 
\begin{figure}[h]
  \centering
  \includegraphics[width=0.6\linewidth]{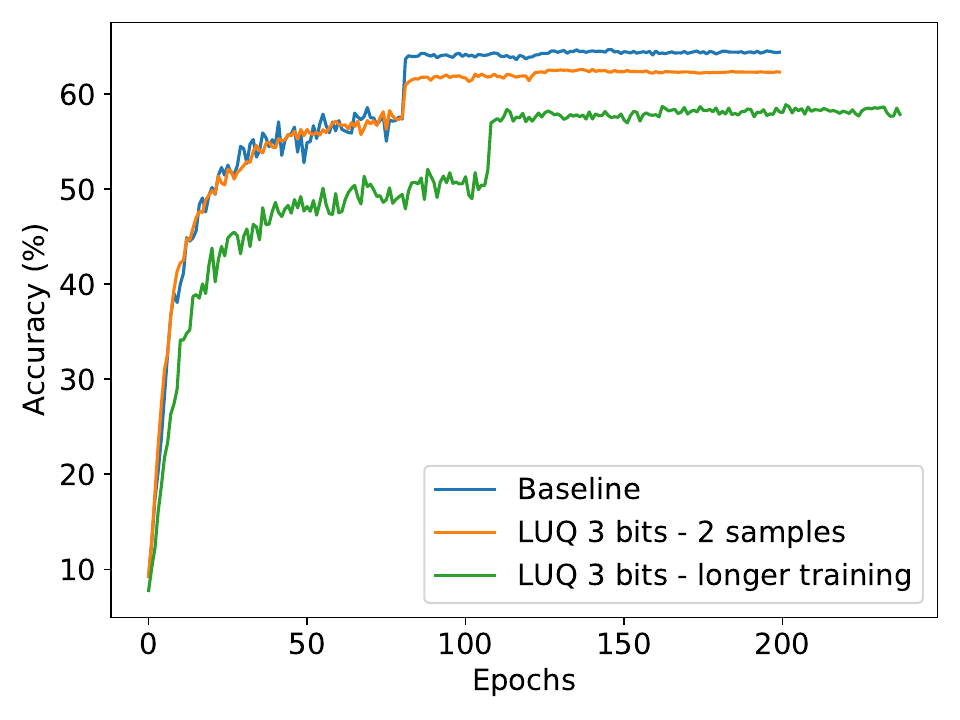}  
  \caption{Comparison of  ResNet-18 3 bit training on Cifar100 dataset of LUQ with 2 samples with longer training of regular LUQ. Both methods have similar overhead, but the SMP method leads to better accuracy.}
    \label{fig:luqLonger}
\end{figure}

The SMP method (\cref{sec:rpt}) has a power overhead of $\sim \frac{1}{3}$ of the number of additional samples since it influences only the update GEMM. In \cref{fig:luqLonger} we compare LUQ with one additional sample which has $\sim 33 \%$ power overhead with regular LUQ with additional $\sim 33 \%$ epochs. The learning rate scheduler was expanded respectively. We can notice that, even though both methods have a similar overhead, the variance reduction achieved with SMP is more important for network accuracy than increasing the training time.

\subsubsection{Data movement reduction effect}
\label{sec:AppHindsight}
\begin{figure*}[h]
\centering
\begin{subfigure}{.45\textwidth}
  \centering
  \includegraphics[width=\linewidth]{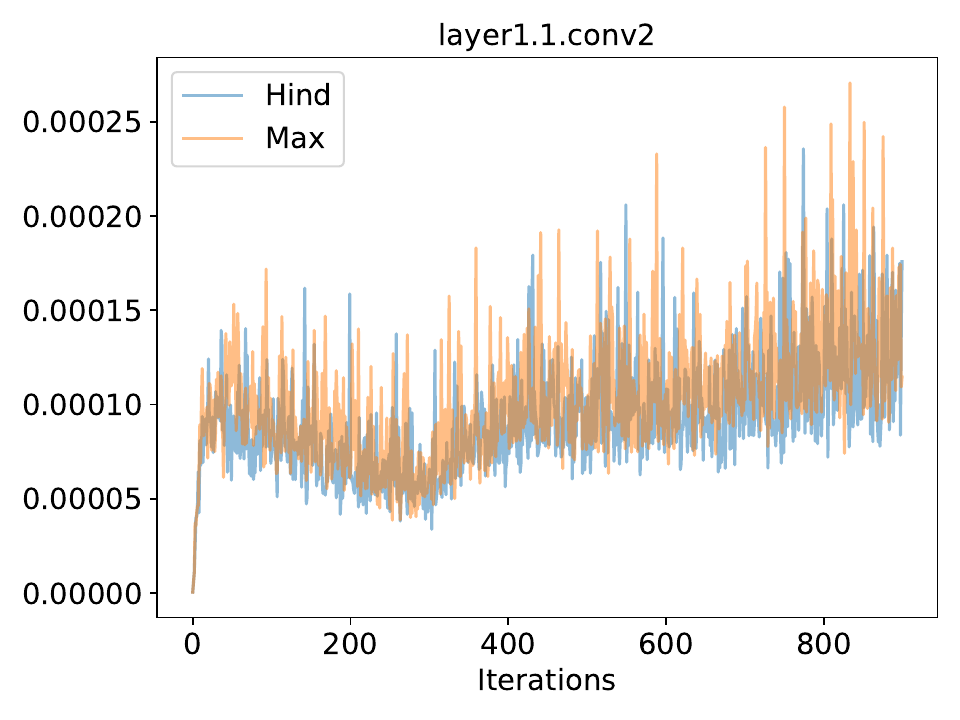} 
  \vspace{-5mm}
  \caption{}
 \end{subfigure}
\begin{subfigure}{.45\textwidth}
  \centering
  \includegraphics[width=\linewidth]{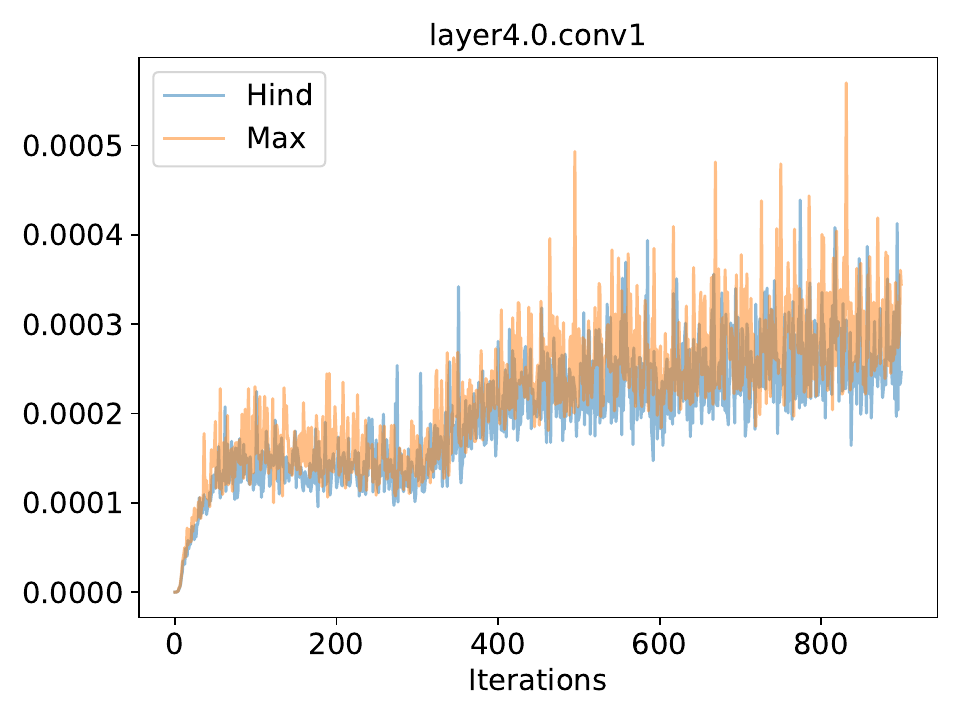}
  \caption{}
\end{subfigure}
\vspace{-3mm}
\caption{Measured maximum and hindsight \cite{Fournarakis2021InHindsightQR} maximum estimation in two different layers of ResNet18 ImageNet dataset. As you can notice, the hindsight estimation is close to the measured maximum and reduce completely the data movement bottleneck} 
\label{fig:hinsightMxMeasure}
\end{figure*}

LUQ requires the measurement of the maximum to choose the underflow threshold $\alpha$ (\cref{sec:luq}). This measurement can create a data movement bottleneck. In order to avoid it, we combine in LUQ the proposed maximum estimation of Hindsight \cite{Fournarakis2021InHindsightQR}, which uses previous iterations statistics. In \cref{fig:hinsightMxMeasure} we compare the measured maximum and the Hindsight estimation, showing they can have similar values. Moreover in \cref{tab:luqWithHindsight} we show that the effect of the Hindsight estimation on the network accuracy is negligible while completely eliminating the data movement bottleneck.  In \cref{tab:expFNTHind} we extend \cref{tab:expFNT} and show the effect of applying the proposed FNT method (\cref{sec:fnt}) on the combination of LUQ with Hindisght.

\begin{table}[h!]
\centering
\caption{Effect of applying Hindsight \cite{Fournarakis2021InHindsightQR} maximum estimation on the network accuracy. SMP refers to doing two samples of the SR quantization of neural gradient in order to reduce the variance as explained in \cref{sec:rpt} similar to \cref{tab:exp} }
\label{tab:luqWithHindsight}
\begin{tabular}{lcccc}
\toprule \multicolumn{1}{l|}{Model} & \multicolumn{1}{c|}{LUQ} & \multicolumn{1}{c|}{LUQ + SMP} & \multicolumn{1}{c|}{LUQ + Hindsight } & \multicolumn{1}{c}{LUQ + Hindsight + SMP }   \\ \midrule
\multicolumn{1}{l|}{ResNet18} & \multicolumn{1}{c|}{69.09} & \multicolumn{1}{c|}{69.24}  & \multicolumn{1}{c|}{69.12} & \multicolumn{1}{c}{69.25}\\ \hline
\multicolumn{1}{l|}{ResNet50} & \multicolumn{1}{c|}{75.42} & \multicolumn{1}{c|}{75.63}  & \multicolumn{1}{c|}{75.4} & \multicolumn{1}{c}{75.6}\\  \hline
\multicolumn{1}{l|}{MobileNet-V2} & \multicolumn{1}{c|}{69.55} & \multicolumn{1}{c|}{69.7} & \multicolumn{1}{c|}{69.59} & \multicolumn{1}{c}{69.68}\\  \hline
\multicolumn{1}{l|}{ResNext50} &  \multicolumn{1}{c|}{76.02} & \multicolumn{1}{c|}{76.12} & \multicolumn{1}{c|}{75.91} & \multicolumn{1}{c}{76.2}\\  \hline
\multicolumn{1}{l|}{Transformer-base} & \multicolumn{1}{c|}{27.17} & \multicolumn{1}{c|}{27.25}  & \multicolumn{1}{c|}{27.17}&\multicolumn{1}{c}{27.23}\\  \hline
\multicolumn{1}{l|}{BERT fine-tune} & \multicolumn{1}{c|}{85.75} & \multicolumn{1}{c|}{85.9}  & \multicolumn{1}{c|}{85.77} & \multicolumn{1}{c}{85.89}\\  
\bottomrule
\end{tabular}
\end{table}

\begin{table*}[h!]
\centering
\caption{Effect of the proposed FNT method (\cref{sec:fnt}) on LUQ and on LUQ + Hindisght \citep{Fournarakis2021InHindsightQR} similar to \cref{tab:expFNT}. FNT 1/2/3 refers to the number of finetuning epochs, while "No FNT" refers to the basic method + SMP as presented in \cref{tab:luqWithHindsight}} 
\label{tab:expFNTHind}
\begin{tabular}{lccccc}
\toprule
\multicolumn{1}{l|}{Model} & \multicolumn{1}{c|}{Baseline} & \multicolumn{1}{c|}{No FNT} & \multicolumn{1}{c|}{+FNT 1} &  \multicolumn{1}{c|}{+FNT 2} &  \multicolumn{1}{c}{+FNT 3} \\ \midrule
\multicolumn{1}{l|}{ResNet-18 LUQ} & \multicolumn{1}{c|}{69.7 \%} & \multicolumn{1}{c|}{69.24 \%} & \multicolumn{1}{c|}{69.7 \%} &  \multicolumn{1}{c|}{-} & 
\multicolumn{1}{c}{-} \\ \hline
\multicolumn{1}{l|}{ResNet-18 LUQ+Hind} & \multicolumn{1}{c|}{69.7 \%} & \multicolumn{1}{c|}{69.25 \%} & \multicolumn{1}{c|}{69.7 \%} &  \multicolumn{1}{c|}{-} & 
\multicolumn{1}{c}{-} \\ \hline
\multicolumn{1}{l|}{ResNet-50 LUQ} & \multicolumn{1}{c|}{76.5 \%} & \multicolumn{1}{c|}{75. 63\%} & \multicolumn{1}{c|}{75.89 \%} &  \multicolumn{1}{c|}{76 \%} & 
\multicolumn{1}{c}{76.18 \%} \\ \hline
\multicolumn{1}{l|}{ResNet-50 LUQ + Hind} & \multicolumn{1}{c|}{76.5 \%} & \multicolumn{1}{c|}{75. 6\%} & \multicolumn{1}{c|}{75.84 \%} &  \multicolumn{1}{c|}{76.03 \%} & 
\multicolumn{1}{c}{76.25 \%} \\ \hline
\multicolumn{1}{l|}{MobileNet-V2 LUQ} & \multicolumn{1}{c|}{71.9 \%} & \multicolumn{1}{c|}{69.7 \%} & \multicolumn{1}{c|}{70.1 \%} &  \multicolumn{1}{c|}{70.3 \%} & 
\multicolumn{1}{c}{70.3 \%} \\ \hline
\multicolumn{1}{l|}{MobileNet-V2 LUQ + Hind} & \multicolumn{1}{c|}{71.9 \%} & \multicolumn{1}{c|}{69.68 \%} & \multicolumn{1}{c|}{70.1 \%} &  \multicolumn{1}{c|}{70.3 \%} & 
\multicolumn{1}{c}{70.3 \%} \\ \hline
\multicolumn{1}{l|}{ResNext-50 LUQ} & \multicolumn{1}{c|}{77.6 \%} & \multicolumn{1}{c|}{76.12\%} & \multicolumn{1}{c|}{76.25 \%} &  \multicolumn{1}{c|}{76.33 \%} & 
\multicolumn{1}{c}{76.7 \%} \\
\hline
\multicolumn{1}{l|}{ResNext-50 LUQ + Hind} & \multicolumn{1}{c|}{77.6 \%} & \multicolumn{1}{c|}{76.2\%} & \multicolumn{1}{c|}{76.3 \%} &  \multicolumn{1}{c|}{76.43 \%} & 
\multicolumn{1}{c}{76.65 \%} \\
\bottomrule
\end{tabular}
\end{table*}

\begin{table}[h!]
\centering
\caption{ResNet-50 accuracy with ImageNet dataset while using quantization on different parts of the network. The forward phase is quantized to INT4 format with SAWB \citep{SAWB} while the backward phase is quantized with the proposed LUQ. As expected, the quantization of the backward phase makes more degradation to the network accuracy.}
\label{tab:FwdvsBwd}
\begin{tabular}{lcc}
\toprule \multicolumn{1}{l|}{Forward} & \multicolumn{1}{c|}{Backward} &  \multicolumn{1}{c}{Accuracy}   \\ \midrule
\multicolumn{1}{l|}{FP32} & \multicolumn{1}{c|}{FP32} & \multicolumn{1}{c}{76.5 \%}\\ \cline{1-3}
\multicolumn{1}{l|}{INT4} & \multicolumn{1}{c|}{FP32} & \multicolumn{1}{c}{76.35 \%}\\ \cline{1-3}
\multicolumn{1}{l|}{FP32} & \multicolumn{1}{c|}{FP4} & \multicolumn{1}{c}{75.6 \%}\\ \cline{1-3}
\multicolumn{1}{l|}{INT4} & \multicolumn{1}{c|}{FP4} & \multicolumn{1}{c}{75.4 \%}
  \\ \bottomrule
\end{tabular}
\end{table}

\subsection{Comparison to \cite{UltraLowP4}}
\label{sec:UltraLowCompare}

Floating point radix conversion requires an explicit multiplication and may require additional non-standard hardware support \citep{Kupriianova_2013}. Specifically, \cite{UltraLowP4} requires a conversion from the radix-2 FP32 to a radix-4 FP4 of the neural gradients. They show this conversion requires multiplication by the constant 1.6. 

Notice that it is not possible to convert between radix floating point formats by a fixed shift. For example, suppose we first convert the radix-2 FP32 to radix-2 FP4 and then shift the exponent (where the shift is equivalent to multiplication by 2). This would lead to an incorrect result, as we show with a simple example: let us assume radix-2 quantization with bins 1,2,4,8 and radix-4 quantizations with bins 1,4,16,64. For the number 4.5, if we quantize it first to radix-2 we get the quantized number 4, then we multiply it by 2 we get 8. In contrast, radix-4 quantization should give the result 4.

In contrast, our proposed method, LUQ, uses the standard radix-2 format and does not require non-standard conversions. The use of standard hardware increases the benefit of the low bits quantization.

\subsection{Power-of-two LUQ}
\label{sec:appHWFriend}

Recall the quantization bins in LUQ are $2^n\alpha$ ($n\in\{0,1,..,b-1\})$, where $\alpha$ is a real number defined as (with $b=3$ for FP4): $$\alpha = \frac{\text{max}(|x|)}{2^{2^{b-1}}}\, .$$
In order to reduce the computational resources in the quantization, we can use only power-of-two values for $\alpha$. This allows us to convert all bins to power-to-two values. This enables using a cheap shift operation instead of the more expensive multiplication with $\alpha$ we do during the quantization process (\cref{eq:unbiasedFpQ}). 

Specifically, we suggest power-of-two LUQ ($\text{LUQ}_{\text{PW2}}$), which uses the ceiling power-of-two value of the maximum in the $\alpha$ calculation, i.e.: $$\alpha_{\mathrm{PW2}} = \frac{2^{\lceil\log_2{\text{max}(|x|)}\rceil}}{2^{2^{b-1}}}\, .$$  The choice of the ceiling (instead of round-to-nearest) is to avoid clipping of the maximum value which will create a bias and affect the accuracy. 

Standard FP quantization, with $E$ exponent bits and $M$ mantissa has a dynamic range of  $[2^{1-q} ,2^{2^E-2-q}(2-2^{-M})]$ where $q$ is the exponent bias. Usually, the exponent bias is fixed as $q=2^{E-1}-1$. However, in modern deep learning accelerator \cite{Tesla} the FP quantizers have the ability to change the exponent bias. With this ability, during the use of $\text{LUQ}_{\text{PW2}}$, we can define the exponent bias for a tensor $x$ to be $q=2^3-2-\lceil \log_2 (\text{max}(|x|)) \rceil$ (for $E=3$ and $M=0$). This bias allows us to completely avoid any shifting operations, since it is defined as part of the quantizer. Thus we obtain an even more significant reduction in the computational resources required to use LUQ.  


In \cref{tab:expHWLUQ} we show results of the proposed $\text{LUQ}_{\text{PW2}}$ achieving a small degradation in comparison to standard LUQ and reducing or completely avoiding (depends on the accelerator) the computational cost of the scaling operator. Moreover in \cref{tab:expHWLUQHind} we combine $\text{LUQ}_{\text{PW2}}$ with Hindsight \cite{Fournarakis2021InHindsightQR} to additionally reduce the data movement, showing better results than \cite{UltraLowP4} with their non hardware friendly method.
\begin{table*}[h!]
\centering
\caption{{Comparison of Ultra-low \citep{UltraLowP4}, LUQ and $\text{LUQ}_{\text{PW2}}$ on various models and datasets. As can be seen, the proposed $\text{LUQ}_{\text{PW2}}$ avoids the scaling operations and achieved a small degradation in comparison to standard LUQ.}}
\label{tab:expHWLUQ}
\begin{tabular}{lcccc}
\toprule
\multicolumn{1}{l|}{Model} & \multicolumn{1}{c|}{Baseline} & \multicolumn{1}{c|}{Ultra-low} & \multicolumn{1}{c|}{$\text{LUQ}$} &  \multicolumn{1}{c}{$\text{LUQ}_\text{PW2}$}  \\ \midrule
\multicolumn{1}{l|}{ResNet-18} & \multicolumn{1}{c|}{69.7 \%} & \multicolumn{1}{c|}{68.27 \%} &  \multicolumn{1}{c|}{69.09\%} &  \multicolumn{1}{c}{{69 \%}}\\ \hline
\multicolumn{1}{l|}{ResNet-50} & \multicolumn{1}{c|}{76.5\%} & \multicolumn{1}{c|}{74.01 \%} &  \multicolumn{1}{l|}{75.42 \%} &    \multicolumn{1}{c}{{75.15 \%}} \\ \hline
\multicolumn{1}{l|}{MobileNet-V2} & \multicolumn{1}{c|}{71.9 \%} & \multicolumn{1}{c|}{68.85 \%} &
\multicolumn{1}{c|}{69.55 \%} &\multicolumn{1}{c}{69.2 \%}\\  \hline
\multicolumn{1}{l|}{ResNext50} & \multicolumn{1}{c|}{77.6 \%} & \multicolumn{1}{c|}{N/A} &
\multicolumn{1}{c|}{76.02 \%} & \multicolumn{1}{c}{75.3 \%}\\  \hline
\multicolumn{1}{l|}{Transformer-base} & \multicolumn{1}{c|}{27.5 (BLEU)} & \multicolumn{1}{c|}{25.4}&
\multicolumn{1}{c|}{27.17} &\multicolumn{1}{c}{26.86}\\  \hline
\multicolumn{1}{l|}{BERT fine-tune} & \multicolumn{1}{c|}{87.03 (F1)} & \multicolumn{1}{c|}{N/A} &
\multicolumn{1}{c|}{85.75} &\multicolumn{1}{c}{85.29}\\  

\bottomrule
\end{tabular}
\end{table*}

\begin{table*}[h!]
\centering
\caption{Combination of the proposed $\text{LUQ}_{\text{PW2}}$ and Hindsight \cite{Fournarakis2021InHindsightQR} on ResNet18 and ResNet50 ImageNet dataset. As can be seen, the Hindsight method has a small effect on the accuracy and allow to reduce the data movement, which can be critical in some cases. }
\label{tab:expHWLUQHind}
\begin{tabular}{lcccc}
\toprule
\multicolumn{1}{l|}{Model} & \multicolumn{1}{c|}{Baseline} & \multicolumn{1}{c|}{Ultra-low} &  \multicolumn{1}{c|}{$\text{LUQ}_\text{PW2}$} &  \multicolumn{1}{c}{$\text{LUQ}_\text{PW2}$ + Hindsight}  \\ \midrule
\multicolumn{1}{l|}{ResNet-18} & \multicolumn{1}{c|}{69.7 \%} & \multicolumn{1}{c|}{68.27 \%}  &  \multicolumn{1}{c|}{{68.7 \%}} &    \multicolumn{1}{c}{{68.88 \%}} \\ \hline
\multicolumn{1}{l|}{ResNet-50} & \multicolumn{1}{c|}{76.5 \%} & \multicolumn{1}{c|}{74.01\%}  & \multicolumn{1}{l|}{75.15 \%} &    \multicolumn{1}{c}{{74.83 \%}}  \\ 
\bottomrule
\end{tabular}
\end{table*}

\subsection{MF-BPROP: multiplication free backpropagation}
\label{sec:mfBprop}
\begin{figure*}[h]
\centering
\begin{subfigure}{.7\textwidth}
  \centering
  \includegraphics[width=\linewidth]{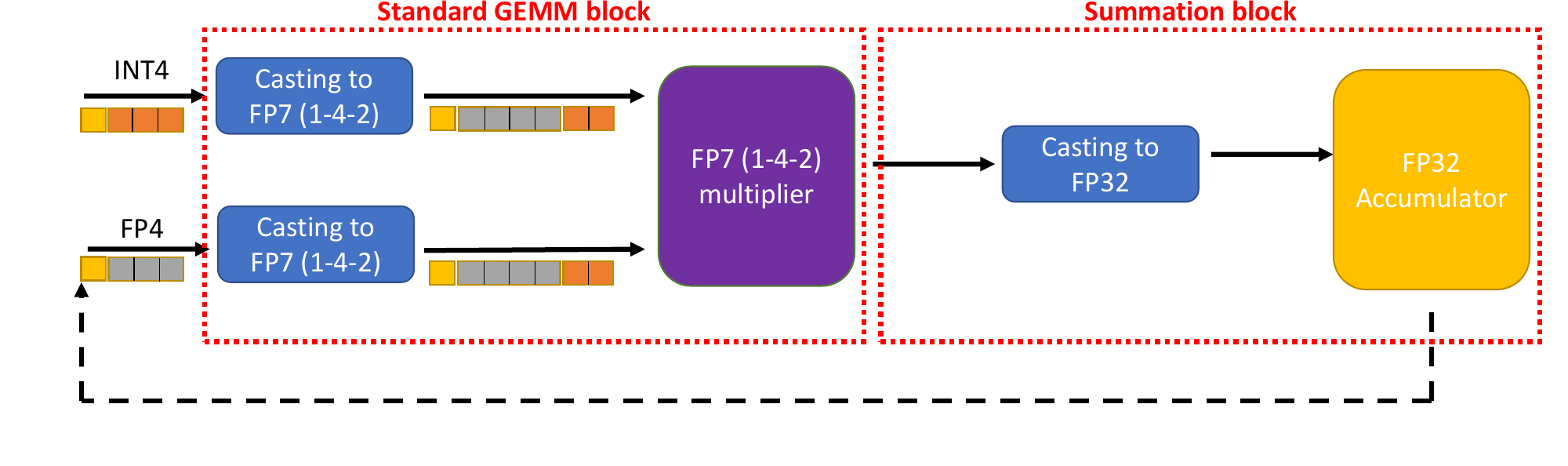} 
  \vspace{-5mm}
  \caption{}
  \label{fig:HwChartBefore}
 \end{subfigure}
\begin{subfigure}{.7\textwidth}
  \centering
  \includegraphics[width=\linewidth]{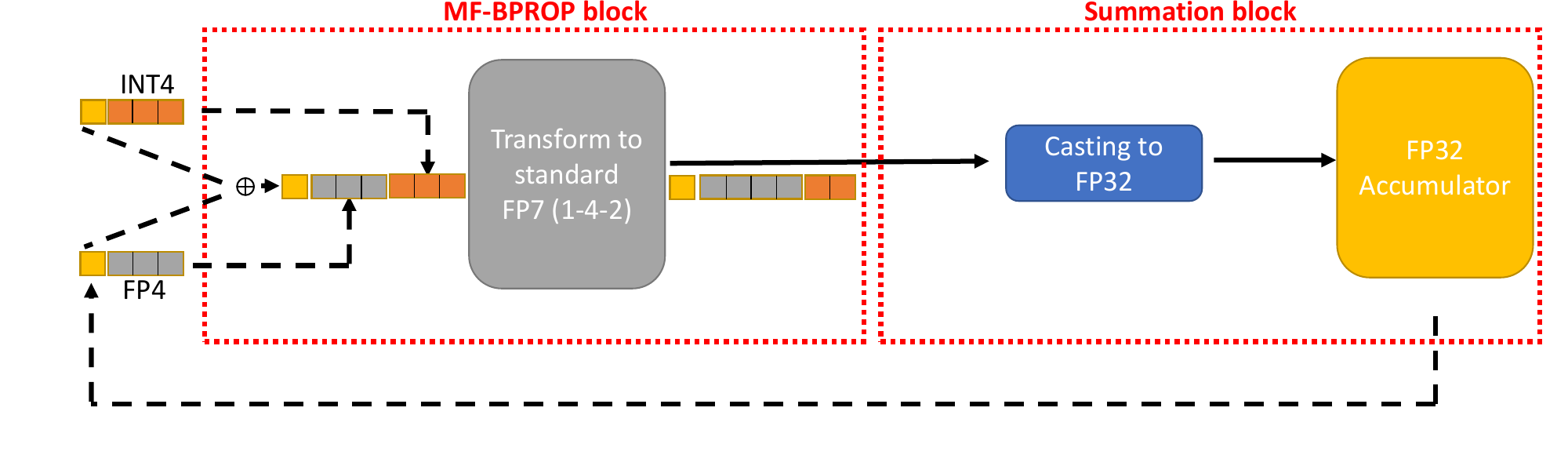} 
  \vspace{-5mm}
  \caption{}
    \label{fig:HwChartAfter}
\end{subfigure}
\vspace{-3mm}
\caption{\textbf{(a):} Standard MAC block illustration containing the two main blocks - one for GEMM and second for accumulator. The GEMM block for hybrid datatype as in our case (FP4 and INT4) requires casting to a common datatype before being inserted into the multiplier. \textbf{(b):} The suggested MAC block, which replaces the multiplier with the proposed MF-BPROP. Instead of doing an expensive casting followed by multiplication, we propose to make only a simple XOR and a transformation (\cref{sec:appTransfromFP7}) reducing the GEMM area by 5x (\cref{sec:appHwAnalysis}). } 
\label{fig:HwFullChart}
\end{figure*}

The main problem of using different datatypes for the weights, activations and neural gradients is the need to cast them to a common data type before the multiplication during the backward (\cref{eq:backward}) and update(\cref{eq:update}) phases.  During the backward and update phases, in each layer $l$ there are two GEMMs between different datatypes:

\vspace{-2mm}

Regularly, to calculate these GEMMs there is a need to cast both data types to a common data type (in our case, FP7 [1,4,2]), then do the GEMM and finally, the results are usually accumulated in a wide accumulator (\cref{fig:HwChartBefore}). This casting cost is not negligible. For example, casting INT4 to FP7 consumes $\sim 15\%$ of the area of an FP7 multiplier. 

In our case, we are dealing with a special case where we do a GEMM between a number without mantissa (neural gradient) and a number without exponent (weights and activations), since INT4 is almost equivalent to FP4 with format [1,0,3]. We suggest transforming the standard GEMM block (\cref{fig:HwChartBefore}) to Multiplication Free BackPROP (MF-BPROP) block which contains only a transformation to standard FP7 format (see \cref{fig:HwChartAfter}) and a simple XOR operation. More details on this transformation appear in \cref{sec:appTransfromFP7}. In our analysis (\cref{sec:appHwAnalysis}) we show the MF-BPROP block reduces the area of the standard GEMM block by $5\times$. Since the FP32 accumulator is still the most expensive block when training with a few bits, we reduce the total area in our experiments by $\sim 8\%$. However, as previously showen \citep{Wang2018TrainingDN}
16-bits accumulators work well with 8-bit training, so it is reasonable to think, it should work also with 4-bit training. In this case, the analysis (\cref{sec:appHwAnalysis}) shows that the suggested MF-BPROP block reduces the total area by $\sim 22\%$.

\subsubsection{Transform to standard fp7}
\label{sec:appTransfromFP7}

\begin{figure}[h]
  \centering
  \includegraphics[width=\linewidth]{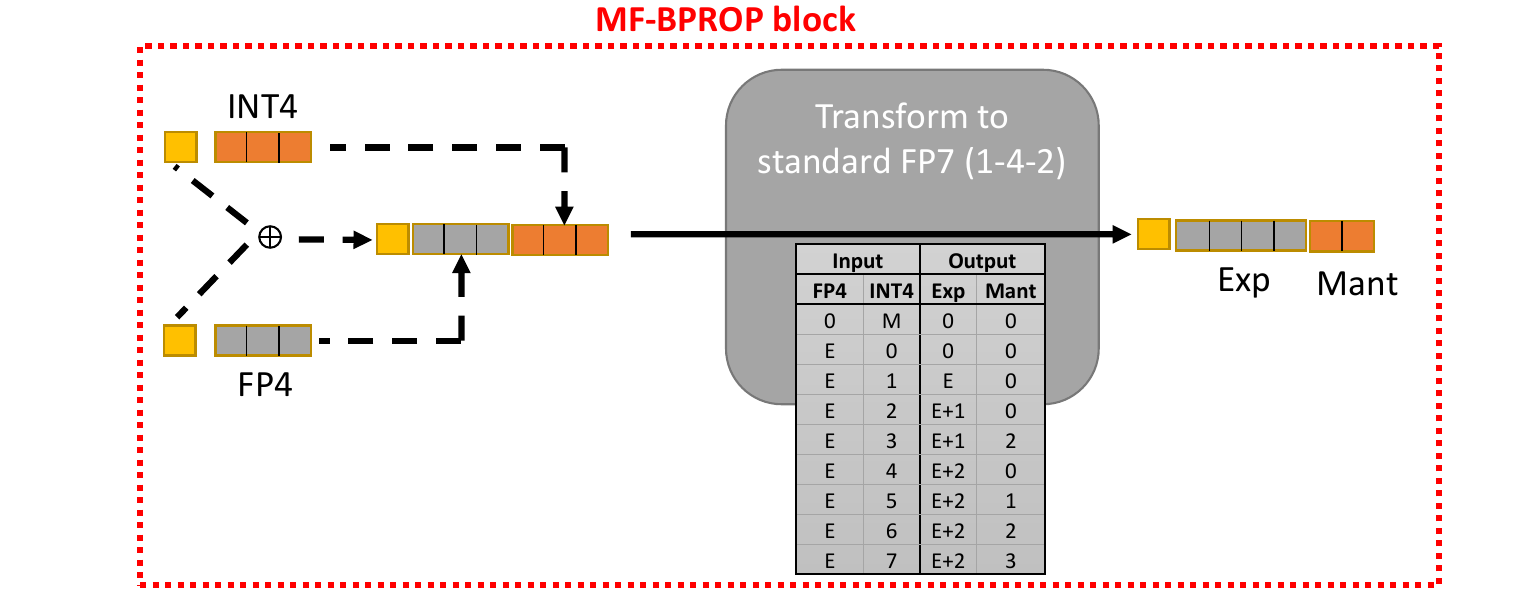}  
  \caption{Illustration of MF-BPROP block which replaces a standard multiplication. It includes: (1) a  simple xor operation between the sign. (2) A transform to standard FP7 format. We present the table to make this transform - E and M represent the bits of the FP4 and INT4 respectively without the sign. Exp and Mant are the bits of the output exponent (4-bit) and mantissa (2-bit) of the output in FP7 format.}
\label{fig:SuggestedMult}
\end{figure}

We suggest a method to avoid the use of an expensive GEMM block between the INT4 (activation or weights) and FP4 (neural gradient).  It includes 2 main elements: The first is a simple xor operation between the sign of the two numbers and the second is a transform block to standard FP7 format. In \cref{fig:SuggestedMult} we present an illustration of the proposed method.  The transformation can be explained with a simple example: for simplicity, we avoid the sign which requires only xor operation. The input arguments are 3 (011 bits representation in INT4 format) and 4 (011 bits representation in FP4 1-3-0 format). The concatenation brings to the bits 011 011. Then looking at the table in the input column where the M=3 (since the INT4 argument = 3) and get the results in FP7 format of 0100 10 ( = E+1 2) which is 12 in FP7 (1-4-2) as the expected multiplication result.

In the next section, we analyze the area of the suggested block in comparison to the standard GEMM block, showing a $5\times$ area reduction.

\subsubsection{Backpropagation without multiplcation analysis}
\label{sec:appHwAnalysis}

In this section, we show a rough estimation of the logical area of the proposed MF-BPROP block which avoids multiplication and compares it with the standard multiplier. In hardware design, the logical area can be a good proxy for power consumption \citep{10.5555/280564}. Our estimation doesn't include synthesis optimization. In \cref{tabApp:multEstimate} we show the estimation of the number of gates of a standard multiplier, getting 264 logical gates while the proposed MF-BPROP block has an estimation of 49 gates (\cref{tabApp:nmtEstimate}) achieving a $\sim 5\times$ area reduction. For fair comparison we remark that in the proposed scheme the FP32 accumulator is the most expensive block with an estimation of 2453 gates, however we believe it can be reduced to a narrow accumulator such as FP16 (As previously shown in \cite{Wang2018TrainingDN} which have an estimated area of 731 gates. In that case, we reduce the total are by $\sim 22 \%$.

\begin{table}[h]
\centering
\caption{Rough estimation of the number of logical gates for a standard GEMM block which contains two blocks: a casting to FP7 and a FP7 multiplier. }
\label{tabApp:multEstimate}
\begin{tabular}{l|l|c}
\toprule
Block & Operation & \# Gates \\ \midrule
\multirow{2}{*}{Casting to FP7} & Exponent 3:1 mux & 12 \\ \cline{2-3} 
 & Mantissa 4:1 mux & 18 \\ \hline
\multirow{6}{*}{FP7 [1,4,2] multiplier} & Mantissa multiplier & 99 \\ \cline{2-3} 
 & Exponent adder & 37 \\ \cline{2-3} 
 & Sign xor & 1 \\ \cline{2-3} 
 & Mantissa normalization & 48 \\ \cline{2-3} 
 & Rounding adder & 12 \\ \cline{2-3} 
 & Fix exponent & 37 \\ \hline
\multicolumn{2}{c|}{\textbf{Total}} & \textbf{264} \\ \bottomrule
\end{tabular}
\end{table}

\begin{table}[h]
\centering
\caption{Rough estimation of the number of logical gates for the proposed MF-BPROP block.}
\label{tabApp:nmtEstimate}
\begin{tabular}{l|l|c}
\toprule
Block & Operation & \# Gates \\ \midrule
\multirow{3}{*}{MF-BPROP} & Exponent adder & 30 \\ \cline{2-3} 
 & Mantissa 4:1 mux & 18 \\ \cline{2-3} 
 & Sign xor & 1 \\ \hline
\multicolumn{2}{c|}{\textbf{Total}} & \textbf{49} \\ \bottomrule
\end{tabular}
\end{table}

\end{document}